\documentclass[10pt,twocolumn,letterpaper]{article}

\usepackage[pagenumbers]{cvpr} 
\usepackage{graphicx}
\usepackage{amsmath}
\usepackage{amssymb}
\usepackage{mathrsfs}
\usepackage{mathtools}
\usepackage{booktabs}
\usepackage{rotating}
\usepackage{multirow}
\usepackage{booktabs}
\usepackage{lipsum}
\usepackage{xcolor}
\usepackage{enumitem}
\usepackage{inconsolata}
\usepackage{pifont}
\usepackage{soul}
\usepackage{balance}
\usepackage{subcaption}
\usepackage[accsupp]{axessibility}
\usepackage[pagebackref,breaklinks,colorlinks]{hyperref}

\usepackage[capitalize]{cleveref}
\crefname{section}{Sec.}{Secs.}
\Crefname{section}{Section}{Sections}
\Crefname{table}{Table}{Tables}
\crefname{table}{Tab.}{Tabs.}

\makeatletter
\def\input@path{{arxiv_sections/}}
\makeatother
\graphicspath{ {./media/} }

\makeatletter
\def\blfootnote{\xdef\@thefnmark{}\@footnotetext}
\makeatother

\newcommand{\modelname}{EmoTx}

\renewcommand{\paragraph}[1]{\vspace{1mm}\noindent\textbf{#1}}
\newcommand{\bbf}{\mathbf{f}}
\newcommand{\bu}{\mathbf{u}}

\newcommand{\bc}{\mathbf{c}}

\newcommand{\by}{\mathbf{y}}
\newcommand{\bz}{\mathbf{z}}

\newcommand{\bW}{\mathbf{W}}
\newcommand{\bE}{\mathbf{E}}

\newcommand{\CLS}{\mathsf{CLS}}
\newcommand{\bbR}{\mathbb{R}}

\newcommand{\MM}{\mathcal{M}}
\newcommand{\MU}{\mathcal{U}}

\newcommand{\MV}{\mathcal{V}}
\newcommand{\MS}{\mathcal{S}}

\newcommand{\MC}{\mathcal{C}}

\newcommand{\MP}{\mathcal{P}}

\newcommand{\ML}{\mathcal{L}}

\newcommand{\dingcheck}{\ding{51}}

\begin{document}
\title{How you feelin'? Learning Emotions and Mental States in Movie Scenes}

\author{
Dhruv Srivastava
\hspace{1cm}
Aditya Kumar Singh
\hspace{1cm}
Makarand Tapaswi\\
CVIT, IIIT Hyderabad, India \\
\small{\url{https://katha-ai.github.io/projects/emotx}}
}
\maketitle

\begin{abstract}

Movie story analysis requires understanding characters' emotions and mental states.
Towards this goal, we formulate emotion understanding as predicting a diverse and \emph{multi-label} set of emotions at the level of a movie scene and for each character.
We propose \modelname{}, a multimodal Transformer-based architecture that ingests videos, multiple characters, and dialog utterances to make joint predictions.
By leveraging annotations from the MovieGraphs dataset~\cite{moviegraphs}, we aim to predict classic emotions (\eg~happy, angry) and other mental states (\eg~honest, helpful).
We conduct experiments on the most frequently occurring 10 and 25 labels, and a mapping that clusters 181 labels to 26.
Ablation studies and comparison against adapted state-of-the-art emotion recognition approaches shows the effectiveness of \modelname{}.
Analyzing \modelname's self-attention scores reveals 
that expressive emotions often look at character tokens while other mental states rely on video and dialog cues.

\end{abstract}
\section{Introduction}
\label{sec:intro}

In the movie \emph{The Pursuit of Happyness}, we see the protagonist
experience a roller-coaster of emotions from the lows of breakup and homelessness to the highs of getting selected for a coveted job.
Such heightened emotions are often useful to draw the audience in through relatable events as one empathizes with the character(s).
For machines to understand such a movie (broadly, story), we argue that it is paramount to track how characters' emotions and mental states evolve over time.
Towards this goal, we leverage annotations from MovieGraphs~\cite{moviegraphs} and train models to watch the video, read the dialog, and predict the emotions and mental states of characters in each movie scene.

\begin{figure}[t]
\centering
\includegraphics[width=\linewidth]{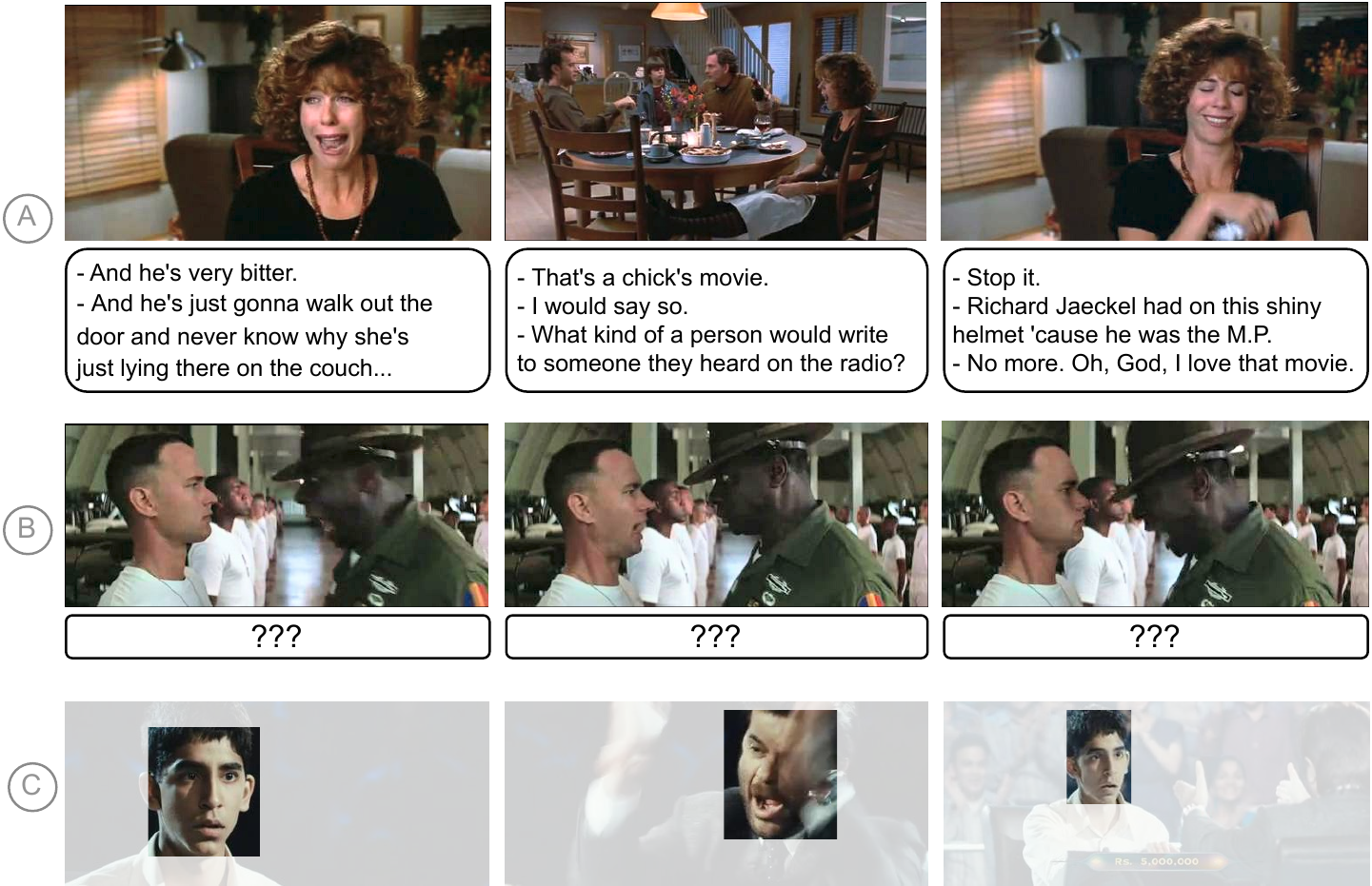}
\vspace{-4mm}
\caption{Multimodal models and multi-label emotions are necessary for understanding the story.
\textbf{A}: What character emotions can we sense in this scene?
Is a single label enough?
\textbf{B}: Without the dialog, can we try to guess the emotions of the Sergeant and the Soldier.
\textbf{C}: Is it possible to infer the emotions from the characters' facial expressions (without subtitles and visual background) only?
Check the footnote below for the ground-truth emotion labels for these scenes and Appendix~\ref{sec:teaser_details} for an explanation of the story.}

\vspace{-4mm}
\label{fig:teaser}
\end{figure}

Emotions are a deeply-studied topic.
From ancient Rome and Cicero's 4-way classification~\cite{cicero-emo}, to modern brain research~\cite{progbrainres}, emotions have fascinated humanity.
Psychologists use of Plutchik's wheel~\cite{plutchik} or the proposal of universality in facial expressions by Ekman~\cite{Ekman1971}, structure has been provided to this field through various theories.
Affective emotions are also grouped into mental (affective, behavioral, and cognitive) or bodily states~\cite{clore1987affectivelexicon}.

A recent work on recognizing emotions with visual context, Emotic~\cite{emotic} identifies 26 label clusters and proposes a \emph{multi-label} setup wherein an image may exhibit multiple emotions (\eg~\emph{peace, engagement}).
An alternative to the categorical space, valence, arousal, and dominance are also used as three continuous dimensions~\cite{emotic}.

Predicting a rich set of emotions requires analyzing multiple contextual modalities~\cite{emotic, caer, emoticon}.
Popular directions in multimodal emotion recognition are
Emotion Recognition in Conversations (ERC) that classifies the emotion for every dialog utterance~\cite{poria2019meld, dialogRNN, todkat};
or predicting a single valence-activity score for short $\sim$10s movie clips~\cite{BaveyeLIRIS, affect2mm}.

\blfootnote{
Ground-truth emotions and mental states portrayed in movie scenes in Fig.~\ref{fig:teaser}:
\textbf{A}: excited, curious, confused, annoyed, alarmed;
\textbf{B}: shocked, confident;
\textbf{C}: happy, excited, amused, shocked, confident, nervous.}

We operate at the level of a \emph{movie scene}: a set of shots telling a sub-story, typically at one location, among a defined cast, and in a short time span of 30-60s.
Thus, scenes are considerably longer than single dialogs~\cite{poria2019meld} or movie clips in~\cite{BaveyeLIRIS}.
We predict emotions and mental states for all characters in the scene and also by accumulating labels at the scene level.
Estimation on a larger time window naturally lends itself to multi-label classification as characters may portray multiple emotions simultaneously (\eg~\emph{curious} and \emph{confused}) or have transitions due to interactions with other characters (\eg~\emph{worried} to \emph{calm}).

We perform experiments with multiple label sets: Top-10 or 25 most frequently occurring emotion labels in MovieGraphs~\cite{moviegraphs} or a mapping to the 26 labels in the Emotic space, created by~\cite{affect2mm}.
While emotions can broadly be considered as part of mental states, for this work, we consider that
\emph{expressed emotions} are apparent by looking at the character, \eg~\emph{surprise, sad, angry}; and
\emph{mental states} are latent and only evident through interactions or dialog, \eg~\emph{polite, determined, confident, helpful}%
\footnote{Admittedly it is not always easy or possible to categorize a label as an expressed emotion or a mental state, \eg~\emph{cheerful, upset}.
Using Clore~\etal~\cite{clore1987affectivelexicon}'s classification,
\emph{expressed emotions} refer to affective and bodily states, while
our \emph{mental states} refer to behavioral and cognitive states.
}.
We posit that classification in a rich label space of emotions requires looking at multimodal context as evident from masking context in Fig.~\ref{fig:teaser}.
To this end, we propose \modelname{} that jointly models video frames, dialog utterances, and character appearance.

We summarize our contributions as follows:
(i)~Building on rich annotations from MovieGraphs~\cite{moviegraphs}, we formulate scene and per-character emotion and mental state classification as a multi-label problem.
(ii)~We propose a multimodal Transformer-based architecture \modelname{} that predicts emotions by ingesting all information relevant to the movie scene.
\modelname{} is also able to capture label co-occurrence and jointly predicts all labels.
(iii)~We adapt several previous works on emotion recognition for this task and show that our approach outperforms them all. 
(iv)~Through analysis of the self-attention mechanism, we show that the model learns to look at relevant modalities at the right time.
Self-attention scores also shed light on our model's treatment of expressive emotions \vs~mental states.

\section{Related Work}
\label{sec:relwork}

We first present work on movie understanding and then dive into visual and multimodal emotion recognition.
\paragraph{Movie understanding}
has evolved over the last few years from person clustering and identification~\cite{buffy, knock_knock, face_body_voice, brown2021corroborative, c1c, nagrani2017sherlock} to analyzing the story.
Scene detection~\cite{local2global_sceneseg, chen2021shotcol, rotman2017osg,  rasheed2003scenedet, tapaswi2014storygraphs},
question-answering~\cite{movieqa, tvqa, yu2018jsfusion},
movie captioning~\cite{lsmdc, yu2017concepts} with names~\cite{fillin},
modeling interactions and/or relationships~\cite{fan2019understanding, marin2019laeo, lirec},
alignment of text and video storylines~\cite{book2movie, book_movie_uoft, graph_movienet}
and even long-form video understanding~\cite{lvu}
have emerged as exciting areas.
Much progress has been made through datasets such as
Condensed Movies~\cite{condensed_movies}, MovieNet~\cite{movienet},
VALUE benchmark (goes beyond movies)~\cite{value_benchmark},
and MovieGraphs~\cite{moviegraphs}.
Building on the annotations from MovieGraphs~\cite{moviegraphs}, we focus on another pillar of story understanding complementary to the above directions: identifying the emotions and mental states of each character and the overall scene in a movie.

\paragraph{Visual emotion recognition}
has relied on face-based recognition of Ekman's 6 classic emotions~\cite{Ekman1971}, and was popularized through datasets such as MMI~\cite{pantic2005}, CK and CK+~\cite{tian2001CK, lucey2010CKplus}.
A decade ago, EmotiW~\cite{DhallEmotiW13},
FER~\cite{fer13}, and
AFEW~\cite{afew} emerged as challenging in-the-wild benchmarks.
At the same time, approaches such as~\cite{Liu2014DeeplyLD, Liu2014FacialER} introduced deep learning to expression recognition achieving good performance.
Breaking away from the above pattern, the Emotic dataset~\cite{emotic} introduced the use of 26 labels for emotion understanding in images while highlighting the importance of context.
Combining face features and context using two-stream CNNs~\cite{caer} or person detections with depth maps~\cite{emoticon} were considered.
Other directions in emotion recognition include estimating valence-arousal (continuous variables) from faces with limited context~\cite{attendaffectnet}, learning representations through webly supervised data to overcome biases~\cite{PandaDE} or improving them further through a joint text-vision embedding space~\cite{WeiEmotionNet}.
Different from the above, our work focuses on multi-label emotions and mental states recognition in movies exploiting multimodal context both at the scene- and character-level.

\paragraph{Multimodal datasets for emotion recognition}
have seen recent adoption.
Acted Facial Expressions in the Wild~\cite{afew} aims to predict emotions from faces, but does not provide any context.
The Stanford Emotional Narratives Dataset~\cite{ongSENDv1} contains participant shared narratives of positive/negative events in their lives.
While multimodal, these are quite different from edited movies and stories that are our focus.
The Multimodal EmotionLines Dataset (MELD)~\cite{poria2019meld} is an example of Emotion Recognition in Conversations (ERC) and attempts to estimate the emotion for every dialog utterance in TV episodes from \emph{Friends}.
Different from MELD, we operate at the time-scale of a cohesive story unit, a movie scene.
Finally, closest to our work, Annotated Creative Commons Emotional DatabasE (LIRIS-ACCEDE)~\cite{BaveyeLIRIS} obtains emotion annotations for short movie clips.
However, the clips are quite small (8-12s) and annotations are obtained in the continuous valence-arousal space.
Different from the above works, we also aim to predict character-level mental states and demonstrate that video and dialog context helps for such labels.

\paragraph{Multimodal emotion recognition methods.}
RNNs have been used since early days for ERC~\cite{wollmer2010textContext, dialogRNN, jiao2020HMN, sivaprasad2018icmr} (often with graph networks~\cite{ghosal2019dialoguegcn, zhang2019erc}) as they allow effective combination of audio, visual, and textual data.
Inspired by recent advances, Transformer architectures are also adopted for ERC~\cite{m2fnet, shen2021dialogXL}.
External knowledge graphs provide useful commonsense information~\cite{cosmic} while topic modeling integrated with Transformers have improved results~\cite{todkat}.
Multi-label prediction has also been attempted by considering a sequence-to-set approach~\cite{zhang2020multilabelEmotionDet}, however that may not scale with number of labels.
While we adopt a Transformer for joint modeling, our goal to predict emotions and mental states for movie scenes and characters is different from ERC.
We adapt some of the above methods and compare against them in our experiments.
Close to our work, the MovieGraphs~\cite{moviegraphs} emotion annotations are used to model changing emotions across the entire movie~\cite{affect2mm}, and for Temporal Emotion Localization~\cite{Li2022EmoLoc}.
However, the former tracks one emotion in each scene, while the latter proposes a different, albeit interesting direction.

\section{Method}
\label{sec:method}

\begin{figure*}[t]
\centering
\includegraphics[width=\textwidth]{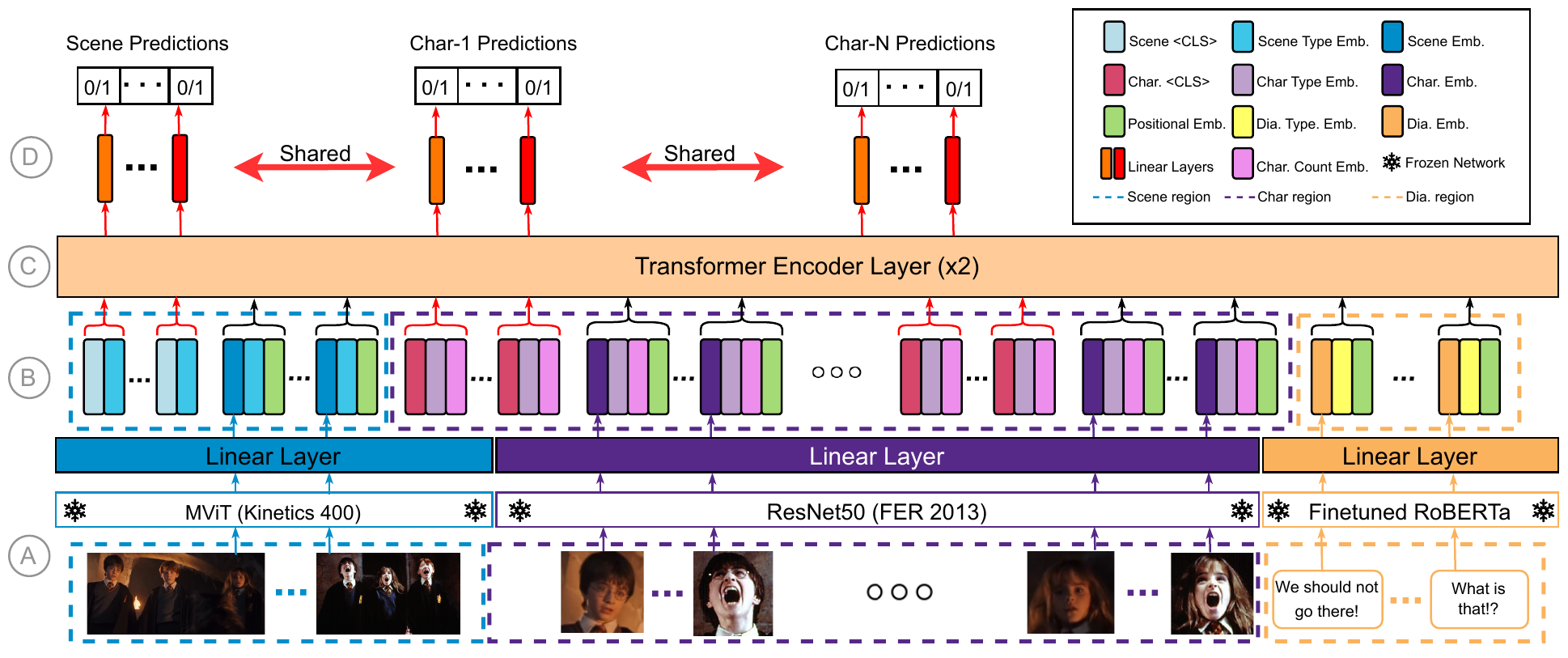}
\vspace{-6mm}
\caption{An overview of \modelname{}.
We present the detailed approach in Sec.~\ref{sec:method} but provide a short summary here.
\textbf{A}: Video features (in blue region),
character face features (in purple region), and
utterance features (in orange region)
are obtained using frozen backbones and projected with linear layers into a joint embedding space.
\textbf{B}: Here appropriate embeddings are added to the tokens to distinguish between modalities, character count, and to provide a sense of time.
We also create per-emotion classifier tokens associated with the scene or a specific character.
\textbf{C}: Two Transformer encoder layers perform self-attention across the sequence of input tokens.
\textbf{D}: Finally, we tap the classifier tokens to produce output probability scores for each emotion through a linear classifier shared across the scene and characters.}
\label{fig:model_overview}
\vspace{-3mm}
\end{figure*}

\modelname{} leverages the self-attention mechanism in Transformers~\cite{attention} to predict emotions and mental states.
We first define the task (Sec.~\ref{subsec:ps}) and then describe our proposed approach (Sec.~\ref{subsec:approach}),
before ending this section with details regarding training and inference (Sec.~\ref{subsec:train}).

\subsection{Problem Statement}
\label{subsec:ps}

We assume that movies have been segmented automatically~\cite{local2global_sceneseg} or with a human-in-the-loop process~\cite{tapaswi2014storygraphs, moviegraphs} into coherent \emph{scenes} that are self-contained and describe a short part of the story.
The focus of this work is on characterizing emotions within a movie scene that are often quite long (30-60s) and may contain several tens of shot changes.

Consider such a movie scene $\MS$ that consists of a set of video frames $\MV$, characters $\MC$, and dialog utterances $\MU$.
Let us denote the set of video frames as $\MV = \{f_t\}_{t=1}^T$, where $T$ is the number of frames after sub-sampling.
Multiple characters often appear in any movie scene.
We model $N$ characters in the scene as $\MC = \{\MP^i\}_{i=1}^{N}$, where each character $\MP^i = \{(f_t, b_t^i)\}$ may appear in some frame $f_t$ of the video at the spatial bounding box $b_t^i$.
We assume that $b_t^i$ is empty if the character $\MP^i$ does not appear at time $t$.
Finally, $\MU = \{u_j\}_{j=1}^{M}$ captures the dialog utterances in the scene.
For this work, we use dialogs directly from subtitles and thus assume that they are unnamed.
While dialogs may be named through subtitle-transcript alignment~\cite{buffy}, scripts are not always available or reliable for movies.

\paragraph{Task formulation.}
Given a movie scene $\MS$ with its video, character, and dialog utterance,
we wish to predict the emotions \emph{and} mental states (referred as labels, or simply emotions) at both the scene, $\by^{\MV}$,
and per-character, $\by^{\MP^i}$, level.
We formulate this as a multi-label classification problem with $K$ labels, \ie~$\by = \{y_k\}_{k=1}^K$.
Each $y_k\ {\in}\ \{0, 1\}$ indicates the absence or presence of the $k^\text{th}$ label in the scene $y_k^\MV$ or portrayed by some character $y_k^{\MP^i}$.
For datasets with character-level annotations, scene-level labels are obtained through a simple logical \texttt{OR} operation, \ie~$\by^{\MV} = \bigoplus_{i=1}^{N}\by^{\MP^i}$.

\subsection{\modelname: Our Approach}
\label{subsec:approach}
We present \modelname{}, our Transformer-based method that recognizes emotions at the movie scene and per-character level.
A preliminary video pre-processing and feature extraction pipeline extracts relevant representations.
Then, a Transformer encoder combines information across modalities.
Finally, we adopt a classification module inspired by previous work on multi-label classification with Transformers~\cite{q2l}.
An overview of the approach is presented in Fig.~\ref{fig:model_overview}.

\paragraph{Preparing multimodal representations.}
Recognizing complex emotions and mental states (\eg~\emph{nervous, determined}) requires going beyond facial expressions to understand the larger context of the story.
To facilitate this, we encode multimodal information through multiple lenses:
(i)~the video is encoded to capture where and what event is happening;
(ii)~we detect, track, cluster, and represent characters based on their face and/or full-body appearance; and
(iii)~we encode the dialog utterances as information complementary to the visual domain.

A pretrained encoder $\phi_{\MV}$ extracts relevant visual information from a single or multiple frames as $\bbf_t = \phi_{\MV}(\{f_t\})$.
Similarly, a pretrained language model $\phi_\MU$ extracts dialog utterance representations as $\bu_j = \phi_{\MU}(u_j)$.
Characters are more involved as we need to first localize them in the appropriate frames.
Given a valid bounding box $b_t^i$ for person $\MP^i$, we extract character features using a backbone pretrained for emotion recognition as $\bc_t^i = \phi_\MC(f_t, b_t^i)$.

\paragraph{Linear projection.}
Token representations in a Transformer often combine the core information (\eg~visual representation) with meta information such as the timestamp through position embeddings (\eg~\cite{videobert}).
We first bring all modalities to the same dimension with linear layers.
Specifically, we project visual representation
$\bbf_t \in \bbR^{D_{\MV}}$ using $\bW_\MV \in \bbR^{D \times D_\MV}$,
utterance representation $\bu_j \in \bbR^{D_{\MU}}$ using $\bW_\MU \in \bbR^{D \times D_\MU}$, and
character representation $\bc_t^i \in \bbR^{D_{\MC}}$ using $\bW_\MC \in \bbR^{D \times D_\MC}$.
We omit linear layer biases for brevity.

\paragraph{Modality embeddings.}
We learn three embedding vectors $\bE^{\MM} \in \bbR^{D \times 3}$ to capture the three modalities corresponding to (1)~video, (2)~characters, and (3)~dialog utterances.
We also assist the model in identifying tokens coming from characters by including a special character count embedding, $\bE^C \in \bbR^{D \times N}$.
Note that the modality and character embeddings do not encode any specific meaning or imposed order (\eg~higher to lower appearance time, names in alphabetical order) - we expect the model to use this only to distinguish one modality/character from the other.

\paragraph{Time embeddings.}
The number of tokens depend on the chosen frame-rate.
To inform the model about relative temporal order across modalities, we adopt a discrete time binning strategy that translates real time (in seconds) to an index.
Thus, video frame/segment and character box representations fed to the Transformer are associated with their relevant time bins.
For an utterance $u_j$, binning is done based on its middle timestamp $t_j$.
We denote the time embeddings as $\bE^T \in \bbR^{D \times \lceil T^* / \tau \rceil}$, where $T^*$ is the maximum scene duration and $\tau$ is the bin step.
For convenience, $\bE^T_t$ selects the embedding using a discretized index $\lceil t / \tau \rceil$.

\paragraph{Classifier tokens.}
Similar to the classic $\CLS$ tokens in Transformer models~\cite{roberta, vit} we use learnable classifier tokens to predict the emotions.
Furthermore, inspired by Query2Label~\cite{q2l}, we use $K$ classifier tokens rather than tapping a single token to generate all outputs (see Fig.~\ref{fig:model_overview}D).
This allows capturing label co-occurrence within the Transformer layers improving performance.
It also enables analysis of per-emotion attention scores providing insights into the model's workings.
In particular, we use $K$ classifier tokens for scene-level predictions (denoted $\bz_k^{\MS}$) and $N \times K$ tokens for character-level predictions (denoted $\bz_k^i$ for character $\MP^i$, one for each character-emotion pair).

\paragraph{Token representations.}
Combining the features with relevant embeddings provides rich information to \modelname.
The token representations for each input group are as follows:
\allowdisplaybreaks
\begin{align}
\text{scene cls. tokens: }
\label{eq:scene_cls_token}
& \tilde{\bz}_k^{\MS} = \bz_k^{\MS} + \bE^{\MM}_1, \\
\label{eq:char_cls_token}
\text{char. cls. tokens: }
& \tilde{\bz}_k^i = \bz_k^i + \bE^{\MM}_2 + \bE^C_i, \\
\label{eq:video_token}
\text{video: }
& \tilde{\bbf}_t = \bW_\MV \bbf_t + \bE^{\MM}_1 + \bE^T_t, \\
\label{eq:char_token}
\text{character box: }
& \tilde{\bc}_t^i = \bW_\MC \bc_t^i + \bE^{\MM}_2 + \bE^C_i + \bE^T_t, \\
\label{eq:utterance_token}
\text{and utterance: }
& \tilde{\bu}_j = \bW_\MU \bu_j + \bE^{\MM}_3 + \bE^T_{t_j} \, .
\end{align}
Fig.~\ref{fig:model_overview}B illustrates this addition of embedding vectors.
We also perform LayerNorm~\cite{layernorm} before feeding the tokens to the Transformer encoder layers, not shown for brevity.

\paragraph{Transformer Self-attention.}
We concatenate and pass all tokens through $H{=}2$ layers of the Transformer encoder that computes self-attention across all modalities~\cite{attention}.
For emotion prediction, we only tap the outputs corresponding to the classification tokens as
\begin{equation}
[\hat{\bz}_k^{\MS}, \hat{\bz}_k^i] = \mathsf{TransformerEncoder} \left( \tilde{\bz}_k^{\MS}, \tilde{\bbf}_t, \tilde{\bz}_k^i, \tilde{\bc}_t^i, \tilde{\bu}_j \right) \, .
\end{equation}
We jointly encode all tokens spanning $\{k\}_1^K, \{i\}_1^N, \{t\}_1^T$, and $\{j\}_1^M$.

\paragraph{Emotion labeling.}
The contextualized representations for the scene $\hat{\bz}_k^{\MS}$ and characters $\hat{\bz}_k^i$ are sent to a shared linear layer $\bW^E \in \bbR^{K \times D}$ for classification.
Finally, the probability estimates through a sigmoid activation $\sigma(\cdot)$ are:
\begin{equation}
\label{eq:predictions}
\hat{y}_k^{\MS} = \sigma( \bW^E_k \hat{\bz}_k^{\MS} ) \,\,
\text{ and } \,\,
\hat{y}_k^i = \sigma( \bW^E_k \hat{\bz}_k^i ), \,\, \forall k, i \, .
\end{equation}

\subsection{Training and Inference}
\label{subsec:train}
\paragraph{Training.}
\modelname{} is trained in an end-to-end fashion with the \emph{BinaryCrossEntropy} (BCE) loss.
To account for the class imbalance we provide weights $\omega_k$ for the positive labels based on inverse of proportions.
The scene and character prediction losses are combined as
\begin{equation}
\ML = \sum_{k=1}^K \texttt{BCE}(\omega_k, y^{\MV}_k, \hat{y}_k^{\MS}) + \sum_{i=1}^N \sum_{k=1}^K \texttt{BCE}(\omega_k, y^{\MP^i}_k, \hat{y}_k^i) \, .
\end{equation}

\paragraph{Inference.}
At test time, we follow the procedure outlined in Sec.~\ref{subsec:approach} and generate emotion label estimates for the entire scene and each character as indicated in Eq.~\ref{eq:predictions}.

\paragraph{Variations.}
As we will see empirically, our model is very versatile and well suited for adding/removing modalities or additional representations by adjusting the width of the Transformer (number of tokens).
It can be easily modified to act as a unimodal architecture that applies only to video or dialog utterances by disregarding other modalities.

\section{Experiments and Discussion}
\label{sec:exp}

We present our experimental setup in Sec.~\ref{subsec:exp:data} before diving into the implementation details in Sec.~\ref{subsec:exp:impl}.
A series of ablation studies motivate the design choices of our model (Sec.~\ref{subsec:exp:abl}) while we compare against the adapted versions of various SoTA models for emotion recognition in Sec.~\ref{subsec:exp:sota}.
Finally, we present some qualitative analysis and discuss how our model switches from facial expressions to video or dialog context depending on the label in Sec.~\ref{subsec:exp:selfattn}.

\begin{figure}[t]
\centering
\includegraphics[width=0.48\linewidth]{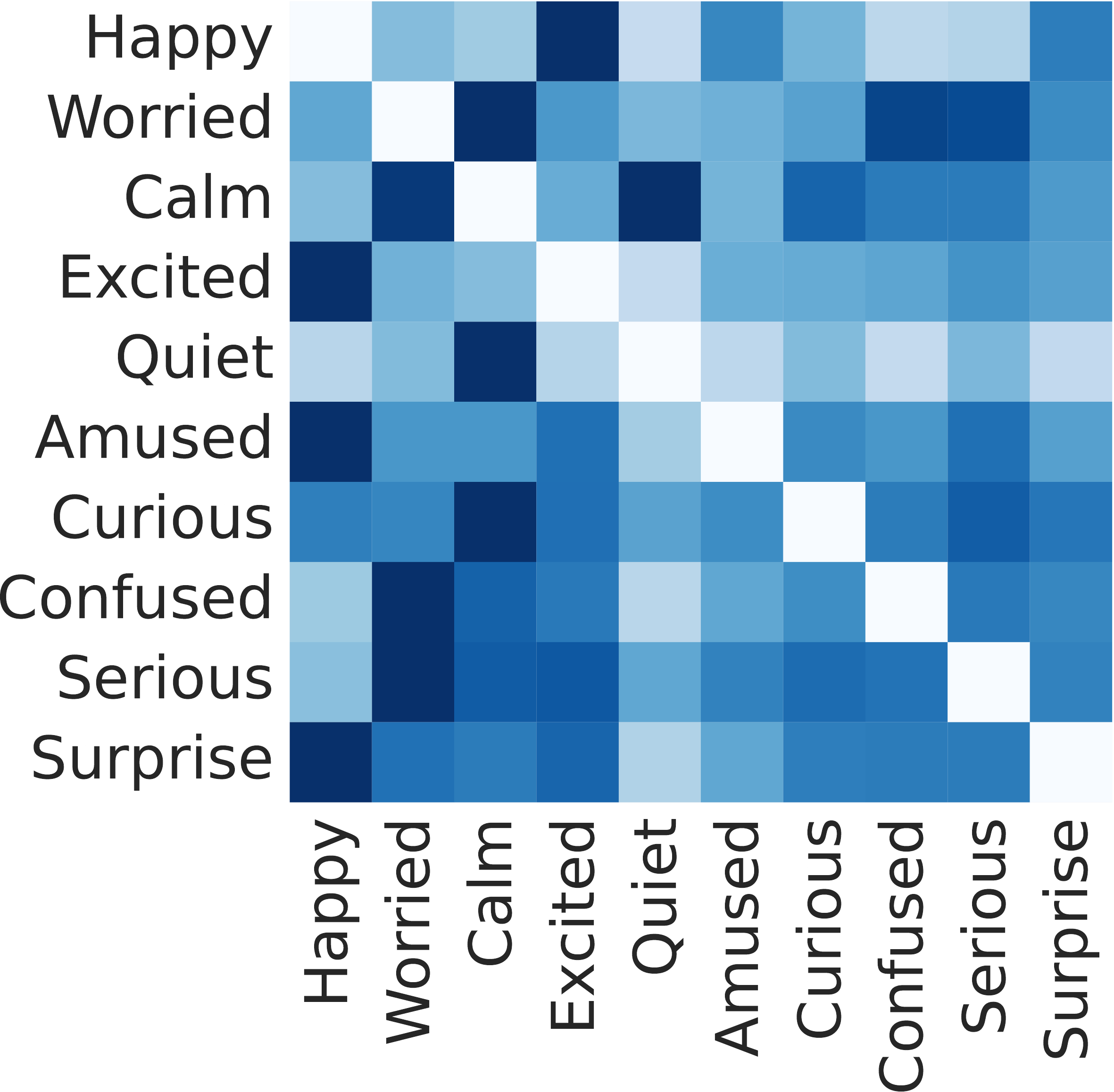}
\includegraphics[width=0.48\linewidth]{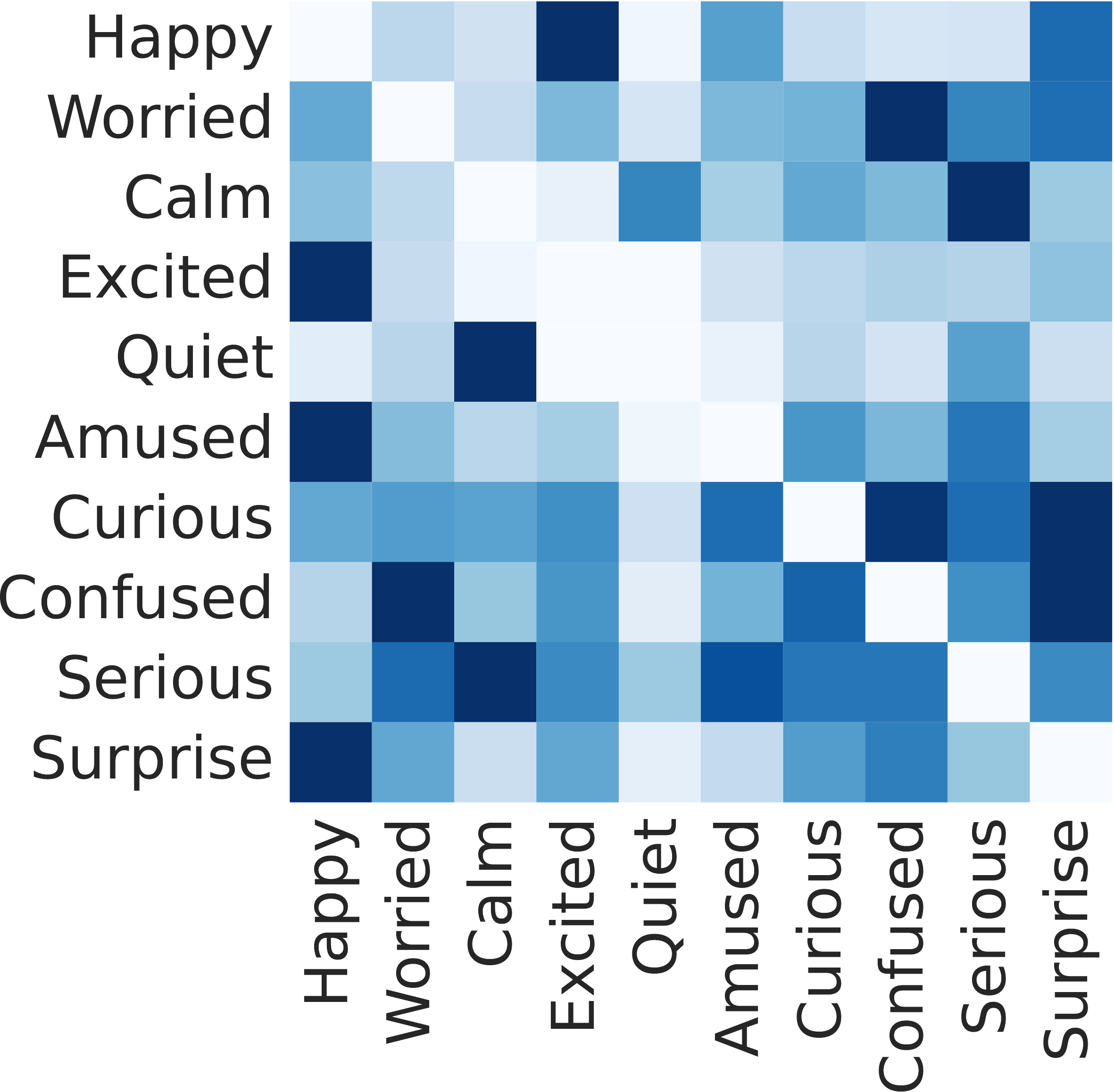}
\vspace{-2mm}
\caption{Row normalized label co-occurrence matrices for the top-10 emotions in a \emph{movie scene} (left) or for a \emph{character} (right).}
\vspace{-4mm}
\label{fig:cooccurrence_maps}
\end{figure}

\begin{figure}[t]
\centering
\includegraphics[width=0.95\linewidth]{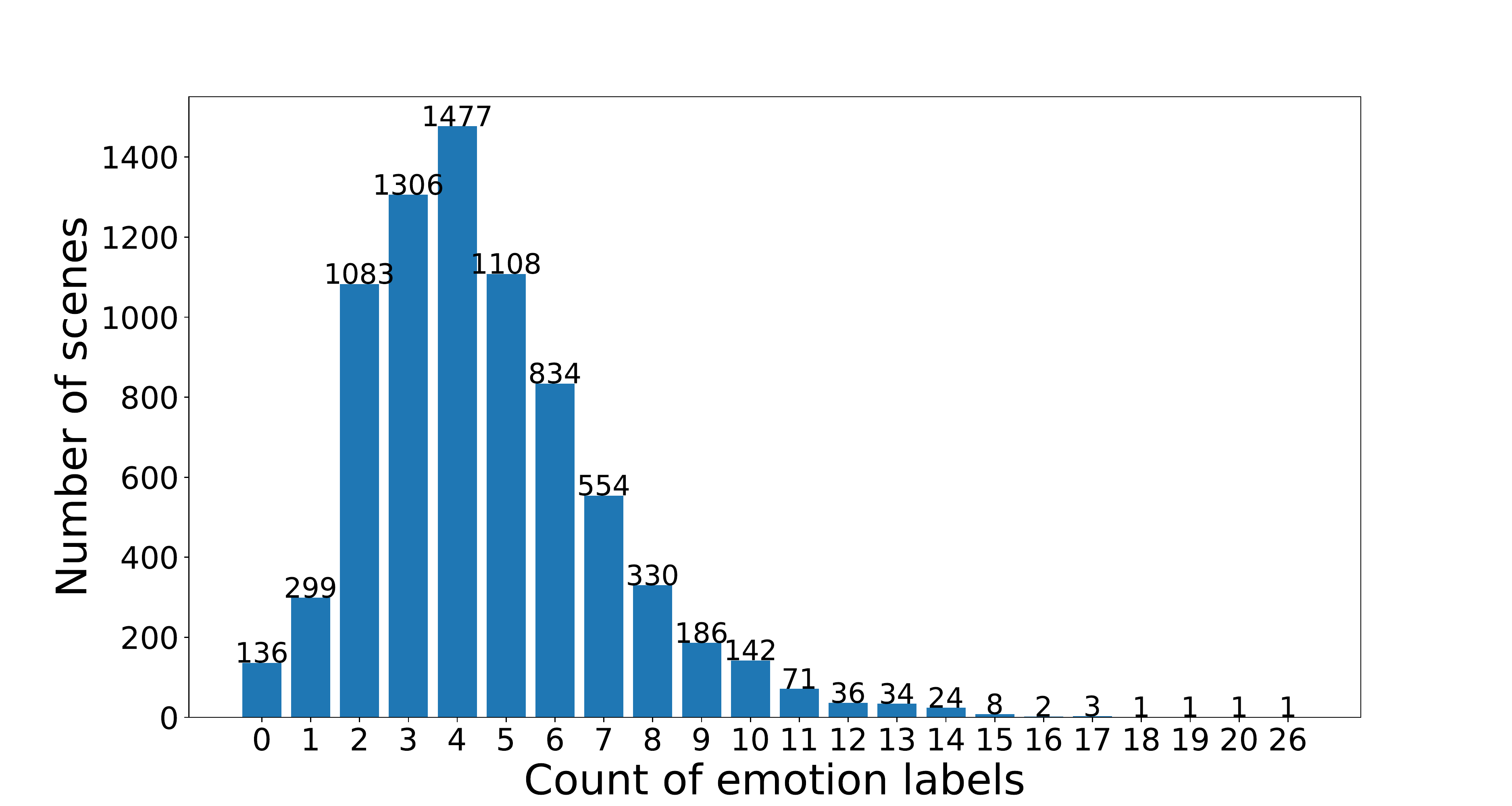}
\vspace{-2mm}
\caption{Bar chart showing the number of movie scenes associated with a specific count of annotated emotions.}
\vspace{-5mm}
\label{fig:emotion_count_distribution}
\end{figure}

\subsection{Dataset and Setup}
\label{subsec:exp:data}
We use the MovieGraphs dataset~\cite{moviegraphs} that features 51 movies and 7637 movie scenes with detailed graph annotations.
We focus on the list of characters and their emotions and mental states, which naturally affords a multi-label setup.
Other annotations such as the situation label, or character interactions and relationships~\cite{lirec} are ignored as they cannot be assumed to be available for a new movie.

\paragraph{Label sets.}
Like other annotations in the MovieGraphs dataset, emotions are also obtained as free-text leading to a huge variability and a long-tail of labels (over 500).
We focus our experiments on three types of label sets:
(i)~\emph{Top-10} considers the most frequently occurring 10 emotions;
(ii)~\emph{Top-25} considers frequently occurring 25 labels; and
(iii)~\emph{Emotic}, a mapping from 181 MovieGraphs emotions to 26 Emotic labels provided by~\cite{affect2mm}.

\paragraph{Statistics.}
We first present row max-normalized co-occurrence matrices for the scene and characters (Fig.~\ref{fig:cooccurrence_maps}).
It is interesting to note how a movie scene has high co-occurrence scores for emotions such as \emph{worried} and \emph{calm} (perhaps owing to multiple characters), while \emph{worried} is most associated with \emph{confused} for a single character.
Another high scoring example for a single character is \emph{curious} and \emph{surprise},
while a movie scene has \emph{curious} with \emph{calm} and \emph{surprise} with \emph{happy}.
In Fig.~\ref{fig:emotion_count_distribution}, we show the number of movie scenes that contain a specified number of emotions.
Most scenes have 4 emotions.
Appendix~\ref{sec:mg_stats} features further analysis.

\paragraph{Evaluation metric.}
We use the original splits from MovieGraphs.
As we have $K$ binary classification problems, we adopt mean Average Precision (mAP) to measure model performance (similar to Atomic Visual Actions~\cite{ava}).
Note that AP also depends on the label frequency.

\subsection{Implementation Details}
\label{subsec:exp:impl}
\paragraph{Feature representations}
\label{subsec:exp:impl:feat_ext}
play a major role on the performance of any model.
We describe different backbones used to extract features for video frames, characters, and dialog.

\underline{Video} features $\bbf_t$:
The visual context is important for understanding emotions~\cite{emotic, caer, emoticon}.
We extract spatial features using ResNet152~\cite{resnet} trained on ImageNet~\cite{imagenet},
ResNet50~\cite{resnet} trained on Place365~\cite{places365}, and
spatio-temporal features, MViT~\cite{FanMViT2021} trained on Kinetics400~\cite{CarreiraQuoVadis2017}.

\underline{Dialo}g features $\bu_j$:
Each utterance is passed through a RoBERTa-Base encoder~\cite{roberta} to obtain an utterance-level embedding.
We also extract features from a RoBERTa model fine-tuned for the task of multi-label emotion classification (based on dialog only).

\underline{Character} features $\bc_t^i$:
are represented based on face or person detections.
We perform face detection with MTCNN~\cite{mtcnn} and person detection with Cascade RCNN~\cite{cascadercnn} trained on MovieNet~\cite{movienet}.
Tracks are obtained using SORT~\cite{sort}, a simple Kalman filter based algorithm, and clusters using C1C~\cite{c1c}.
Details of the character processing pipeline are presented in Appendix~\ref{sec:char_track_extension}. 
ResNet50~\cite{facefeat} trained on SFEW~\cite{sfew} and pretrained on FER13~\cite{fer13} and VGGFace~\cite{vggface},
VGGm~\cite{facefeat} trained on FER13 and pretrained on VGGFace, and InceptionResnetV1~\cite{SzegedyInceptionNet2015} trained on VGGFace2~\cite{CaoVGGF22018} are used to extract face representations.

\paragraph{Frame sampling strategy.}
We sample up to $T {=} 300$ tokens at 3 fps (100s) for the video modality.
This covers $\sim$99\%
of all movie scenes.
Our time embedding bins are also at 3 per second, \ie~$\tau {=} 1/3$s.
During inference, a fixed set of frames are chosen, while during training, frames are randomly sampled from 3 fps intervals which acts as data augmentation.
Character tokens are treated in a similar fashion, however are subject to the character appearing in the video.

\paragraph{Architecture details.}
We experiment with the number of encoder layers, $H\ {\in}\ \{1, 2, 4, 8\}$, but find $H {=} 2$ to work best (perhaps due to the limited size of the dataset).
Both the layers have same configuration - 8 attention heads with hidden dimension of 512.
The maximum number of characters is $N{=}4$ as it covers up to 91\% of the scenes.
Tokens are padded to create batches and to accommodate shorter video clips.
Appropriate masking prevents self-attention on padded tokens.
Put together, \modelname{} encoder looks at $K$ scene classification tokens, $T$ video tokens, $N \cdot (K + T)$ character tokens, and $T$ utterance tokens.
For $K{=}25, N{=}4$ (Top-25 label set), this is up to 1925 padded tokens.

\paragraph{Training details.}
Our model is implemented in PyTorch~\cite{pytorch} and trained on a single NVIDIA GeForce RTX-2080 Ti GPU for a maximum of 50 epochs with a batch size of 8.
The hyperparameters are tuned to achieve best performance on validation set.
We adopt the Adam optimizer~\cite{adam} with an initial learning rate of $5 \times 10^{-5}$, reduced by a factor of 10 using the learning rate scheduler \texttt{ReduceLROnPlateau}.
The best checkpoint maximizes the geometric mean of scene and character mAP.

\begin{figure*}[t]
\centering
\includegraphics[width=0.99\linewidth,height=5cm]{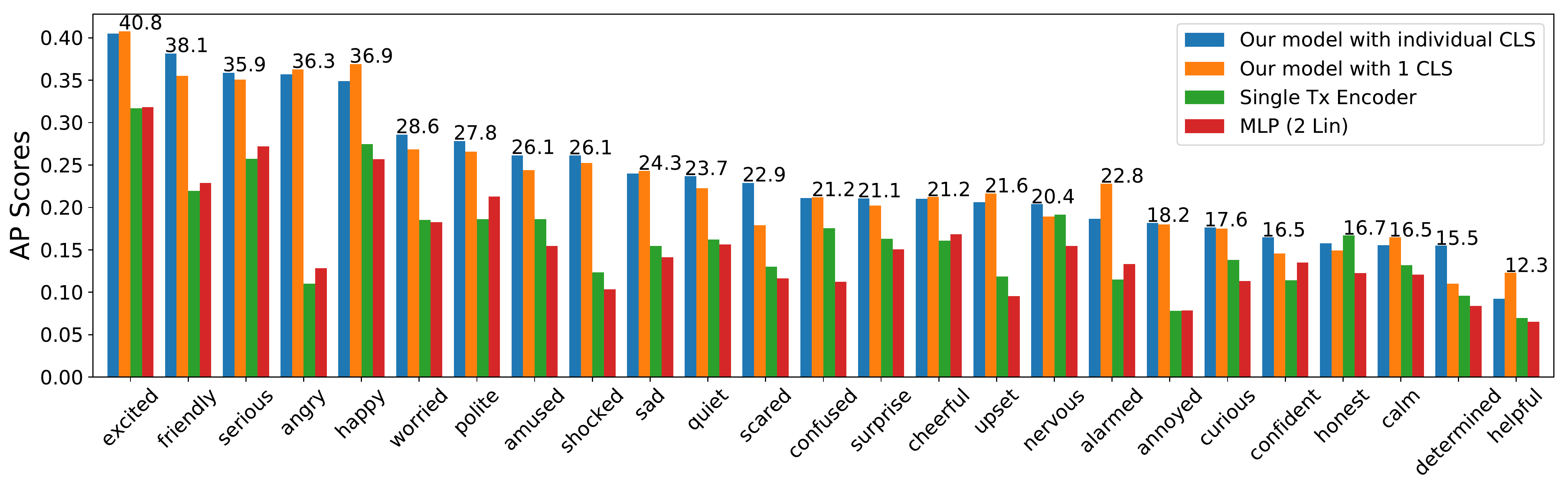}
\vspace{-5mm}
\caption{Comparing scene-level per class AP of \modelname{} against baselines (Table~\ref{tab:arch_abl}) shows consistent improvements.
We also see that our model with $K$ classifier tokens outperforms the 1 CLS token on most classes.
AP of the best model is indicated above the bar.
Interestingly, the order in which emotions are presented is not the same as the frequency of occurrence (see Appendix~\ref{sec:mg_stats}).}
\vspace{-2mm}
\label{fig:per_class_ap}
\end{figure*}

\subsection{Ablation Studies}
\label{subsec:exp:abl}
We perform ablations across three main dimensions: architectures, modalities, and feature backbones.
When not mentioned, we adopt the defaults:
(i)~MViT trained on Kinetics400 dataset to represent video;
(ii)~ResNet50 trained on SFEW, FER, and VGGFace for character representations;
(iii)~fine-tuned RoBERTa for dialog utterance representations; and
(iv)~\modelname{} with appropriate masking to pick modalities or change the number of classifier tokens.

\begin{table}[t]
\centering
\small
\tabcolsep=0.06cm
\begin{tabular}{l cc cc}
\toprule
\multicolumn{1}{c}{\multirow{2}{*}{Method}} & \multicolumn{2}{c}{Top-10} & \multicolumn{2}{c}{Top-25} \\
& Scene & Char  & Scene & Char  \\
\midrule
Random            & 16.87\scriptsize{$\pm$0.23} & 12.49\scriptsize{$\pm$0.15} & 9.73\scriptsize{$\pm$0.101} & 5.84\scriptsize{$\pm$0.05}  \\
\midrule
MLP (2 Lin)       & 23.94\scriptsize{{$\pm$}0.03} & 20.39\scriptsize{$\pm$0.01} & 15.26\scriptsize{{$\pm$}0.02} & 10.57\scriptsize{$\pm$0.02} \\
Single Tx encoder & 25.66\scriptsize{{$\pm$}0.02} & 20.95\scriptsize{$\pm$0.09} & 16.14\scriptsize{{$\pm$}0.03} & 11.08\scriptsize{$\pm$0.18} \\
\midrule
\modelname: 1 CLS & \emph{34.11}\scriptsize{$\pm$0.34} & \emph{23.81}\scriptsize{$\pm$0.24} & \emph{23.34}\scriptsize{$\pm$0.11} & 12.86\scriptsize{$\pm$0.11} \\
\modelname{} (Ours) & \textbf{34.22}\scriptsize{$\pm$0.18} & \textbf{24.35}\scriptsize{$\pm$0.23} & \textbf{23.86}\scriptsize{$\pm$0.10} & \textbf{13.36}\scriptsize{$\pm$0.11} \\
\bottomrule
\end{tabular}
\vspace{-2mm}
\caption{Architecture ablation.
Emotions are predicted at both movie scene and individual character (Char) levels.
We see that our multimodal model significantly outperforms simpler baselines.
Best numbers in bold, close second in italics.}
\vspace{-4mm}
\label{tab:arch_abl}
\end{table}

\paragraph{Architecture ablations.}
We compare our architecture against simpler variants in Table~\ref{tab:arch_abl}.
The first row sets the expectation by providing scores for a \emph{random} baseline
that samples label probabilities from a uniform random distribution between $[0, 1]$ with 100 trials.
Next, we evaluate \emph{MLP (2 Lin)}, a simple MLP with two linear layers with inputs as max pooled scene or character features.
An alternative to max pooling is self-attention.
The \emph{Single Tx encoder} performs self-attention over features (as tokens) and a classifier token to which a multi-label classifier is attached.
Both these approaches are significantly better than random, especially for individual character level predictions which are naturally more challenging than scene-level predictions.

Finally, we compare multimodal \modelname{} that uses 1 classifier token to predict all labels (\modelname: 1 CLS) against $K$ classifier tokens (last row).
Both models achieve significant improvements, \eg~in absolute points, +8.5\% for Top-10 scene labels and +2.3\% for the much harder Top-25 character level labels.
We believe the improvements reflect \modelname{}'s ability to encode multiple modalities in a meaningful way.
Additionally, the variant with $K$ classifier tokens (last row) shows small but consistent +0.5\% improvements over 1 classifier token on Top-25 emotions.

Fig.~\ref{fig:per_class_ap} shows the scene-level AP scores for the Top-25 labels.
Our model outperforms the MLP and Single Tx encoder on 24 of 25 labels and outperforms the single classifier token variant on 15 of 25 labels.
\modelname{} is good at recognizing expressive emotions such as \emph{excited, serious, happy} and even mental states such as \emph{friendly, polite, worried}.
However, other mental states such as \emph{determined} or \emph{helpful} are challenging.

\begin{table}[t]
\centering
\small
\tabcolsep=0.11cm
\begin{tabular}{l cccc cccc}
\toprule
&\multirow{2}{*}{$V_r$} & \multirow{2}{*}{$V_m$} & \multirow{2}{*}{$D$} & \multirow{2}{*}{$C$} & \multicolumn{2}{c}{Top 10 (mAP)} & \multicolumn{2}{c}{Top 25 (mAP)} \\
&           &            &            &             & Scene           & Char           & Scene           & Char           \\
\midrule
1&\dingcheck &      -     &    -       &      -     & 22.81{\tiny $\pm$0.02} & 15.90{\tiny $\pm$0.19}& 14.85{\tiny $\pm$0.02} & 7.98{\tiny $\pm$0.05} \\
2&    -      & \dingcheck &    -       &      -     & 25.73{\tiny $\pm$0.02} & 17.88{\tiny $\pm$0.12}& 16.11{\tiny $\pm$0.05} & 8.96{\tiny $\pm$0.12} \\
3&    -      &      -     & \dingcheck &      -     & 27.28{\tiny $\pm$0.01} & 20.25{\tiny $\pm$0.14}& 20.20{\tiny $\pm$0.08} & 11.09{\tiny $\pm$0.12}\\
4&    -      &      -     &    -       & \dingcheck & 31.38{\tiny $\pm$0.40} & 21.22{\tiny $\pm$0.50}& 20.32{\tiny $\pm$0.05} & 11.23{\tiny $\pm$0.14}\\
\midrule
5&\dingcheck &      -     & \dingcheck &      -     & 27.19{\tiny $\pm$0.07} & 19.45{\tiny $\pm$0.10}& 19.72{\tiny $\pm$0.03} & 10.67{\tiny $\pm$0.08}\\
6&    -      & \dingcheck & \dingcheck &      -     & 28.93{\tiny $\pm$0.02} & 21.41{\tiny $\pm$0.15}& 21.29{\tiny $\pm$0.05} & 12.03{\tiny $\pm$0.23}\\  
7&    -      &      -     & \dingcheck & \dingcheck & 33.59{\tiny $\pm$0.10} & 23.54{\tiny $\pm$0.16}& 23.40{\tiny $\pm$0.09} & 13.01{\tiny $\pm$0.08}\\  
\midrule
8&\dingcheck &      -     & \dingcheck & \dingcheck & 33.60{\tiny $\pm$0.02} & 22.89{\tiny $\pm$0.02}& 22.76{\tiny $\pm$0.02} & 12.21{\tiny $\pm$0.02}\\
9&    -      & \dingcheck & \dingcheck & \dingcheck & \textbf{34.22} {\tiny $\pm$0.18} & \textbf{24.35}{\tiny $\pm$0.23}& \textbf{23.86}{\tiny $\pm$0.10} & \textbf{13.36}{\tiny $\pm$0.11}\\  
\bottomrule
\end{tabular}
\vspace{-2mm}
\caption{
Modality ablation.
$V_{r}$: ResNet50 (Places365),
$V_{m}$: MViT (Kinetics400),
$D$: Dialog, and
$C$: Character.}
\vspace{-4mm}
\label{tab:modal_abl}
\end{table}

\paragraph{Modality ablations.}
We evaluate the impact of each modality (video, characters, and utterances) on scene- and character-level emotion prediction in Table~\ref{tab:modal_abl}.
We observe that the character modality (row 4, R4) outperforms any of the video or dialog modalities (R1-R3).
Similarly, dialog features (R3) are better than video features (R1, R2), common in movie understanding tasks~\cite{moviegraphs, movieqa}.
The choice of visual features is important.
Scene features $V_r$ are consistently worse than action features $V_m$ as reflected in comparisons R1, R2 or R5, R6 or R8, R9.
Finally, we observe that using all modalities (R9) outperforms other combinations, indicating that emotion recognition is a multimodal task.

\begin{figure*}[t]
\centering
\includegraphics[width=0.99\linewidth]{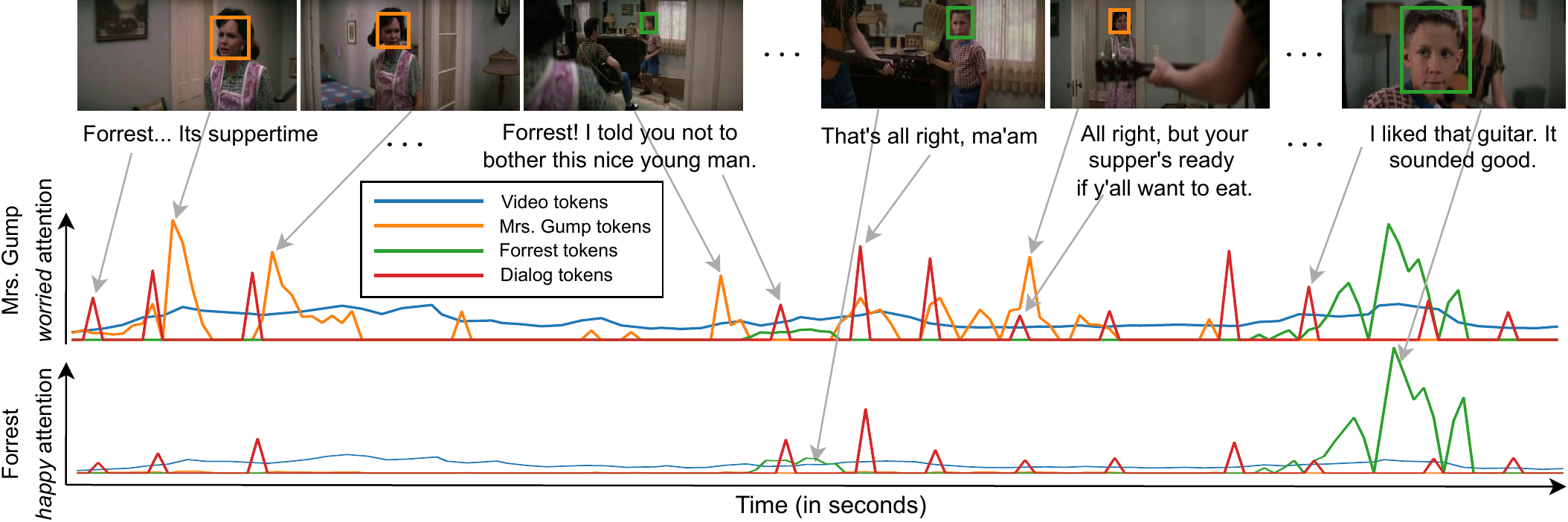}
\vspace{-2mm}
\caption{A scene from the movie \emph{Forrest Gump} showing the multimodal self-attention scores for the two predictions: \emph{Mrs. Gump} is \emph{worried} and \emph{Forrest} is \emph{happy}.
We observe that the \emph{worried} classifier token attends to \emph{Mrs. Gump}'s character tokens when she appears at the start of the scene, while \emph{Forrest}'s \emph{happy} classifier token attends to \emph{Forrest} towards the end of the scene.
The video frames have relatively similar attention scores while dialog helps with emotional utterances such as \emph{told you not to bother} or \emph{it sounded good}.}
\vspace{-4mm}
\label{fig:qualitative_example}
\end{figure*}

\begin{table}[t]
\centering
\tabcolsep=0.06cm
\small
\begin{tabular}{l ccccc cc cc}
\toprule
& \multicolumn{2}{c}{Video} & \multicolumn{2}{c}{Character} &
Dialog & \multicolumn{4}{c}{Metrics} \\
& MViT & R50  & R50 & VGG-M & RB & \multicolumn{2}{c}{Top-10} & \multicolumn{2}{c}{Top-25} \\
& K400 & P365 & FER & FER & FT & Scene & Char & Scene & Char \\
\midrule
1 & - & \dingcheck & - & \dingcheck & No
    & 29.30 & 19.73 & 19.05 & 10.31 \\
2 & \dingcheck & - & - & \dingcheck & No
    & 29.34 & 20.50 & 19.07 & 10.34 \\
3 & - & \dingcheck & \dingcheck & - & No
    & 29.69 & 20.25 & 20.16 & 11.06 \\
4 & \dingcheck & - & \dingcheck & - & No
    & 31.39 & 21.12 & 20.88 & 11.46 \\
5 & \dingcheck & - & - & \dingcheck & \dingcheck
    & 31.50 & 21.60 & 21.49 & 11.64 \\
6 & - & \dingcheck & - & \dingcheck & \dingcheck
    & 32.42 & 22.32 & 21.45 & 11.62 \\
7 & - & \dingcheck & \dingcheck & - & \dingcheck
    & 33.46 & 22.98 & 22.69 & 12.48 \\
8 & \dingcheck & - & \dingcheck & - & \dingcheck
    & \textbf{34.22} & \textbf{24.35} & \textbf{23.86} & \textbf{13.36} \\
    
\bottomrule
\end{tabular}
\vspace{-2mm}
\caption{Feature ablations with backbones.
(MViT, K400): MViT on Kinetics400, 
(R50, P365): ResNet50 on Places365, 
(R50, FER): ResNet50 on Facial Expression Recognition (FER),
(VGG-M, FER): VGG-M on FER, and
(RB, FT): RoBERTa finetuned.
Best numbers in bold.
More results in Appendix~\ref{sec:feature_abl}.}
\vspace{-3mm}
\label{tab:feat_abl}
\end{table}

\paragraph{Backbone ablations.}
We compare several backbones for the task of emotion recognition.
The effectiveness of the fine-tuned RoBERTa model is evident by comparing pairs of rows R2, R5 and R3, R7 and R4, R8 of Table~\ref{tab:feat_abl}, where we see a consistent improvement of 1-3\%.
Character representations with ResNet50-FER show improvement over VGGm-FER as seen from R5, R8 or R6, R7.
Finally, comparing R8 shows the benefits provided by action features as compared to places.
Details are presented in Appendix~\ref{sec:feature_abl}.

\subsection{SoTA Comparison}
\label{subsec:exp:sota}

\begin{table}[t]
\centering
\small
\tabcolsep=0.12cm
\begin{tabular}{l cc cc cc}
\toprule
\multicolumn{1}{c}{\multirow{2}{*}{Method}} & \multicolumn{2}{c}{Top 10}      & \multicolumn{2}{c}{Top 25}      & \multicolumn{2}{c}{Emotic}      \\
\multicolumn{1}{c}{} & Val   & Test  & Val   & Test  & Val   & Test  \\ \midrule
Random               & 16.87 & 13.84 & 9.73 & 7.57 & 11.47 & 11.36 \\
CAER~\cite{caer}
& 18.35 & 15.38 & 11.84 & 9.49  & 13.91 & 12.68 \\
ENet~\cite{WeiEmotionNet}
& 19.14 & 16.14 & 11.22 & 9.08  & 13.55 & 12.64 \\
AANet~\cite{attendaffectnet}
& 21.55 & 17.55 & 12.55 & 10.20 & 14.71 & 13.37 \\
M2Fnet~\cite{m2fnet}
& 24.55 & 19.10 & 16.02 & 13.05 & 18.27 & 16.76 \\
\midrule
\modelname{} (Ours)  & \textbf{34.22} & \textbf{29.35} & \textbf{23.86} & \textbf{19.47} & \textbf{23.67} & \textbf{21.40} \\ \bottomrule
\end{tabular}
\vspace{-3mm}
\caption{Comparison against SoTA for scene-level predictions.
\emph{AANet}: AttendAffectNet.
\emph{ENet}: EmotionNet.
Mean over 3 runs.}
\vspace{-2mm}
\label{tab:sota_scene_abl}
\end{table}

\begin{table}[t]
\centering
\small
\tabcolsep=0.12cm
\begin{tabular}{l cc cc cc}
\toprule
\multicolumn{1}{c}{\multirow{2}{*}{Method}} & \multicolumn{2}{c}{Top 10}      & \multicolumn{2}{c}{Top 25}      & \multicolumn{2}{c}{Emotic}      \\
\multicolumn{1}{c}{} & Val   & Test  & Val   & Test & Val   & Test \\
\midrule
Random
& 12.49 & 11.37 & 5.84 & 5.36 & 6.40 & 6.32 \\
AANet~\cite{attendaffectnet}
& 17.43 & 16.04 & 8.64 & 7.20 & 8.53 & 7.75 \\
M2Fnet~\cite{m2fnet}
& 20.82 & 19.01 & 10.67  & 9.71 & 11.30  & 9.92 \\
\midrule
\modelname{} (Ours)
& \textbf{24.35} & \textbf{22.32} & \textbf{13.36} & \textbf{11.71} & \textbf{12.29} & \textbf{11.76} \\ \bottomrule
\end{tabular}
\vspace{-2mm}
\caption{Comparison against SoTA for character-level predictions.
\emph{AANet} denotes AttendAffectNet.
Mean over 3 runs.}
\vspace{-5mm}
\label{tab:char_sota_abl}
\end{table}

We compare our model against published works 
EmotionNet~\cite{WeiEmotionNet},
CAER~\cite{caer}, 
AttendAffectNet~\cite{attendaffectnet}, and
M2Fnet~\cite{m2fnet}
by adapting them for our tasks (adaptation details are provided in Appendix~\ref{sec:sota_adaptations}).

Table~\ref{tab:sota_scene_abl} shows scene-level performance while the character-level performance is presented in Table~\ref{tab:char_sota_abl}.
First, we note that the test set seems to be harder than val as also indicated by the random baseline, leading to a performance drop from val to test across all approaches.
\modelname{} outperforms all previous baselines by a healthy margin.
For scene level, we see +4.6\% improvement on Emotic labels, +7.8\% on Top-25, and +9.7\% on Top-10.
Character-level predictions are more challenging, but we see consistent improvements of +1.5-3\% across all label sets.
Matching expectation, we see that simpler models such as EmotionNet or CAER perform worse than Transformer-based approaches of M2Fnet and AttendAffectNet.
Note that EmotionNet and CAER are challenging to adapt for character-level predictions and are not presented, but we expect M2Fnet or AttendAffectNet to outperform them.

\subsection{Analyzing Self-attention Scores}
\label{subsec:exp:selfattn}
\modelname{} provides an intuitive way to understand which modalities are used to make predictions.
We refer to the self-attention scores matrix as $\alpha$, and analyze specific rows and columns.
Separating the $K$ classifier tokens allows us to find attention-score based evidence for each predicted emotion by looking at a row $\alpha_{\bz_k^{\MS}}$ in the matrix.

Fig.~\ref{fig:qualitative_example} shows an example movie scene where \modelname{} predicts that \emph{Forrest} is \emph{happy} and \emph{Mrs. Gump} is \emph{worried}.
We see that the model pays attention to the appropriate moments and modalities to make the right predictions.

\begin{figure}[t]
\centering
\includegraphics[width=\linewidth]{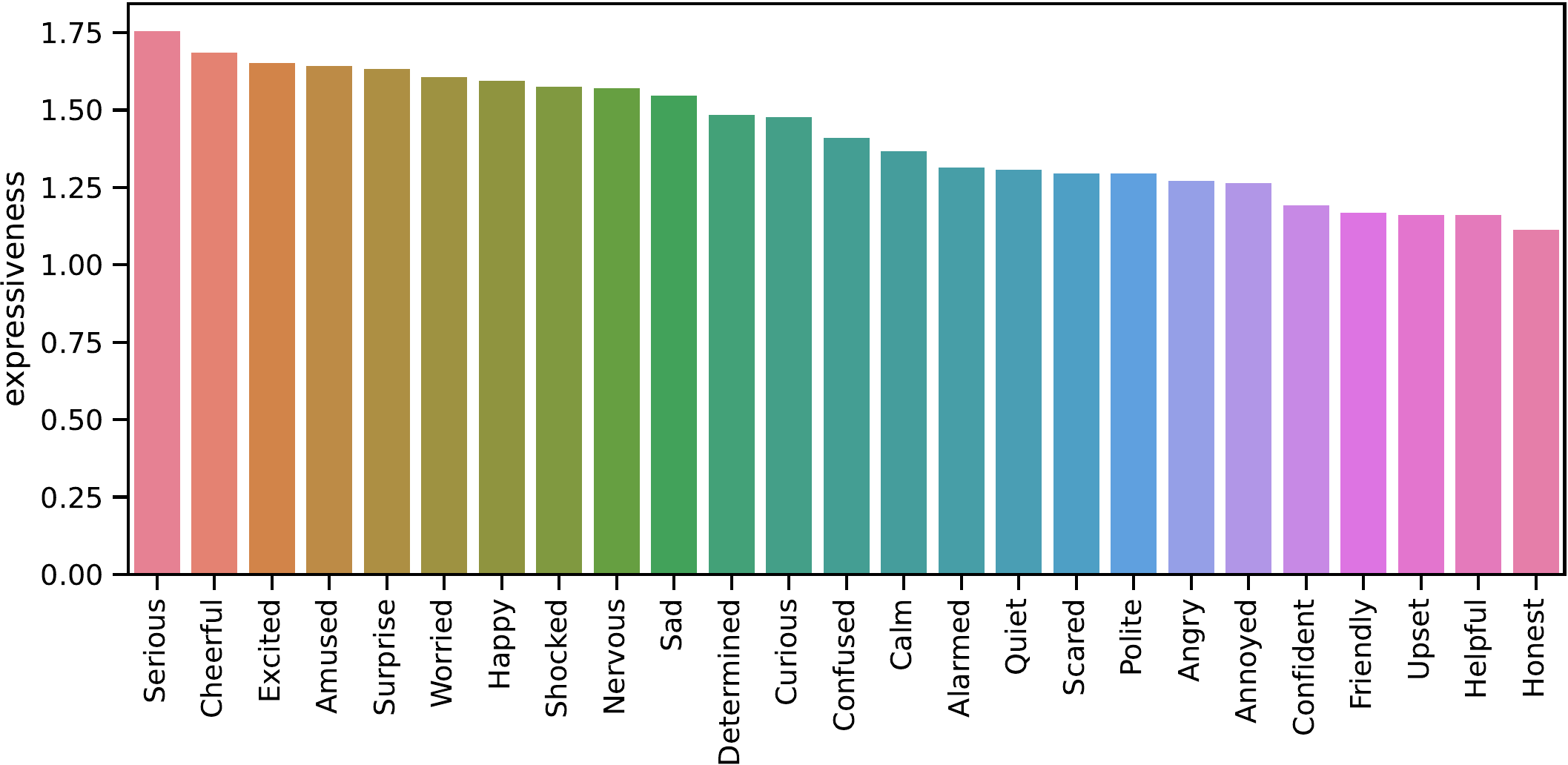}
\vspace{-6mm}
\caption{Sorted expressiveness scores for Top-25 emotions. 
Expressive emotions have higher scores indicating that the model attends to character representations, while mental states have lower scores suggesting more  attention to video and dialog context.}
\vspace{-4mm}
\label{fig:expression}
\end{figure}
\paragraph{Expressive emotions \vs~Mental states.}
We hypothesize that the self-attention module may focus on character tokens for expressive emotions, while looking at the overall video frames and dialog for the more abstract mental states.
We propose an \emph{expressiveness} score as
\begin{equation}
e_k = \dfrac{ \sum_{i=1}^{N} \sum_{t=1}^{T} \alpha_{\bz_k^{\MS}, \bc_t^i} }
{ \sum_{t=1}^{T} \alpha_{\bz_k^{\MS}, \bbf_t} + \sum_{j=1}^{M} \alpha_{\bz_k^{\MS}, \bu_j} } \, ,
\end{equation}
where $\alpha_{\bz_k^{\MS}, \bc_t^i}$ is the self-attention score between the scene classifier token for emotion $k$ ($\bz_k^{\MS}$) and character $\MP^i$'s appearance in the video frame as $b_t^i$;
$\alpha_{\bz_k^{\MS}, \bbf_t}$ is for the video $f_t$ and
$\alpha_{\bz_k^{\MS}, \bu_j}$ is for dialog utterance $u_j$.
Higher scores indicate expressive emotions as the model focuses on the character features, while lower scores identify mental states that analyze the video and dialog context.
Fig.~\ref{fig:expression} shows the averaged expressiveness score for the Top-25 emotions when the emotion is present in the scene (\ie~$y_k{=}1$).
We observe that mental states such as \emph{honest, helpful, friendly, confident} appear towards the latter half of this plot while most expressive emotions such as \emph{cheerful, excited, serious, surprise} appear in the first half.
Note that the expressiveness scores in our work are for faces and applicable to our particular dataset.
We also conduct a short human evaluation to understand expressiveness by annotating whether the emotion is conveyed through video, dialog, or character appearance; presented in Appendix~\ref{sec:supp_qualitative}.

\section{Conclusion}
\label{sec:con}

We presented a novel task for multi-label emotion and mental state recognition at the level of a movie scene and for each character.
A Transformer encoder based model, \modelname{}, was proposed that jointly attended to all modalities (features) and obtained significant improvements over previous works adapted for this task.
Our learned model was shown to have interpretable attention scores across modalities -- they focused on the video or dialog context for mental states while looking at characters for expressive emotions.
In the future, \modelname{} may benefit from audio features or by considering the larger context of the movies instead of treating every scene independently.

\paragraph{Acknowledgements.}
We thank Google India Faculty Research Award 2022 for travel support.

\appendix
\section*{Appendix}

In Sec.~\ref{sec:teaser_details} we refer to Fig.\ref{fig:teaser} (teaser) and share the hidden contexts in each scene reflecting upon the importance of individual modalities to capture the emotions in real-world environments.
Sec.~\ref{sec:mg_stats} present some statistics around emotions extracted from the MovieGraphs dataset.
In Sec.~\ref{sec:char_track_extension} we share the character detection, tracking, and clustering pipeline used to extend the tracks provided in the MovieGraphs dataset.
In Sec.~\ref{sec:AP_analysis} we visualize the class AP scores for top-10 and 25 emotions from MovieGraphs along with Emotic mapped emotions.
Since there were several feature combinations in our work, an extended feature ablation is presented in Sec.~\ref{sec:feature_abl}.
Finally, Sec.~\ref{sec:sota_adaptations} shares details of the modifications made to adapt EmotionNet~\cite{WeiEmotionNet},
CAER~\cite{caer},
M2Fnet~\cite{m2fnet}, and
AttendAffectNet~\cite{attendaffectnet}
for comparison with \modelname.
We end with another qualitative example showing the attention scores similar to Fig.~\ref{fig:qualitative_example} in Sec.~\ref{sec:supp_qualitative}.

\section{The Stories behind Emotions in Fig. 1}
\label{sec:teaser_details}

We discuss some additional details from Fig.\ref{fig:teaser}.
Prior to this, note that the emotions are grouped into three tuples, each corresponding to the frame depicted in the example - however, this was for illustrative purposes and making it easy to match emotions to the frames.
We do not explicitly generate frame-level predictions.

\paragraph{Scene A} is taken from the movie ``Sleepless in Seattle, 1993", scene number 087, where Suzy is narrating an incident from a classical movie ``An affair to remember".
While narrating, she gets sentimental and starts crying. The other characters, Sam and Greg listen curiously but feel neutral and mock her by faking a cry and narrating the scene from some war movie.
This makes Suzy laugh, and she asks the duo to stop before the scene ends.
The reflected emotions and mental states include \emph{upset}, \emph{calm}, \emph{confused}, \emph{excited}, \emph{sad}, and \emph{happy}.
Observing the situation, it is evident that a single emotion label does not suffice and both the visual and dialog context taken over a longer duration is important to predict emotions with mental states.

\paragraph{Scene B}
is taken from the movie ``Forrest Gump, 1994", scene number 045.
Forrest has joined the army and it is his introductory day.
Sergeant Drill asks Forrest about his role in the army to which Forrest replies ``To do whatever you tell me Sergeant Drill" which impresses him a lot.
Then Sergeant Drill praises him by saying it is the best response he has ever heard!
The original subtitles of this clip are shared in Table~\ref{tab:subtitles045}.
We hope to show that the dialog modality is crucial in understanding the real emotions since visually it appears that both the characters are angry and screaming at each other but in reality Forrest is \emph{determined}, \emph{honest}, and \emph{serious}, while the Sergeant is \emph{excited}.

\paragraph{Scene C}
is taken from movie ``Slumdog Millionaire, 2008", scene number 076.
The scene represents a Television Show ``Who wants to be a millionaire?" where Jamal is being asked some question.
He has given the response and is waiting for the confirmation from the anchor. The frames used in the figure reflect the moment when the anchor excitedly reveals that the answer given by Jamal is correct.
However, by only looking at the faces, it appears as if Jamal is tense and he anchor is scolding him, whereas in reality, everyone is clapping and cheering for him.
We show that looking at the visual frame is necessary to correctly predict the wider perspective of emotions, here corresponding to the transition from \emph{nervous} and \emph{curious} to \emph{surprised} and \emph{amused} for Jamal, and \emph{excited} for the anchor.

\begin{table}[b]
\tabcolsep=0.12cm
\small
\begin{tabular}{lllp{4.2cm}}
\toprule
Start & End   & Speaker        & Utterance                                                            \\
\midrule
00:00 & 00:04 & Sergeant Drill & Gump! What's your sole purpose in this Army?                         \\
00:04 & 00:06 & Forrest Gump   & To do whatever you tell me, Drill Sergeant!                          \\
00:06 & 00:10 & Sergeant Drill & God damn it, Gump! You're a goddamn genius!                          \\
00:10 & 00:12 & Sergeant Drill & That's the most outstanding answer I've ever heard.                  \\
00:12 & 00:15 & Sergeant Drill & You must have a goddamn IQ of 160!                                   \\
00:15 & 00:18 & Sergeant Drill & You are goddamn gifted, Private Gump!                                \\
00:19 & 00:21 & Sergeant Drill & Listen up, people!                                                   \\
00:21 & 00:25 & Forrest Gump   & Now, for some reason, I fit in the Army like one of them round pegs. \\
00:25 & 00:27 & Forrest Gump   & It's not really hard.                                                \\
00:27 & 00:30 & Forrest Gump   & You just make your bed real neat and remember to stand up straight,  \\
00:31 & 00:34 & Forrest Gump   & and always answer every question with, "Yes, Drill Sergeant!"        \\
00:35 & 00:36 & Sergeant Drill & Is that clear?                                                       \\
00:36 & 00:38 & Everyone       & Yes, Drill Sergeant!                                                 \\
\bottomrule
\end{tabular}
\caption{Subtitles from Scene 045 from movie Forrest Gump, 1993, corresponding to Scene B from Fig.~1.
Note that the speaker names are added for improving the clarity and understanding, our model does not have access to them.}
\label{tab:subtitles045}
\end{table}

\section{MovieGraphs-Emotions: Dataset Features}
\label{sec:mg_stats}

The MovieGraphs dataset~\cite{moviegraphs} contains graph-based annotations for each scene within a movie.
The nodes of these graphs include characters and their details such as relationships, interactions, emotions, and other physical attributes, along with movie scene-level labels such as the overarching situation, place (scene), and a few sentence natural language description.
There are a total of 51 movies divided into 7637 clips with associated graphs.
The MovieGraphs dataset is provided with train, validation and test splits which contain 33/7/11 movies with 5050/1060/1527 clip graphs respectively.
These clips have an average duration of 41.7s at 23.976 fps (frames per second).
For each clip, we focus on characters and their emotion attributes.
As the dataset consists of free-text annotations, this amounts to massive 509 unique emotion labels in the dataset, which however, can be mapped to a smaller set.

\paragraph{Label distributions.}
We analyze the dataset from various perspectives and highlight some statistics.

\begin{figure}[t]
\centering
\includegraphics[width=\linewidth]{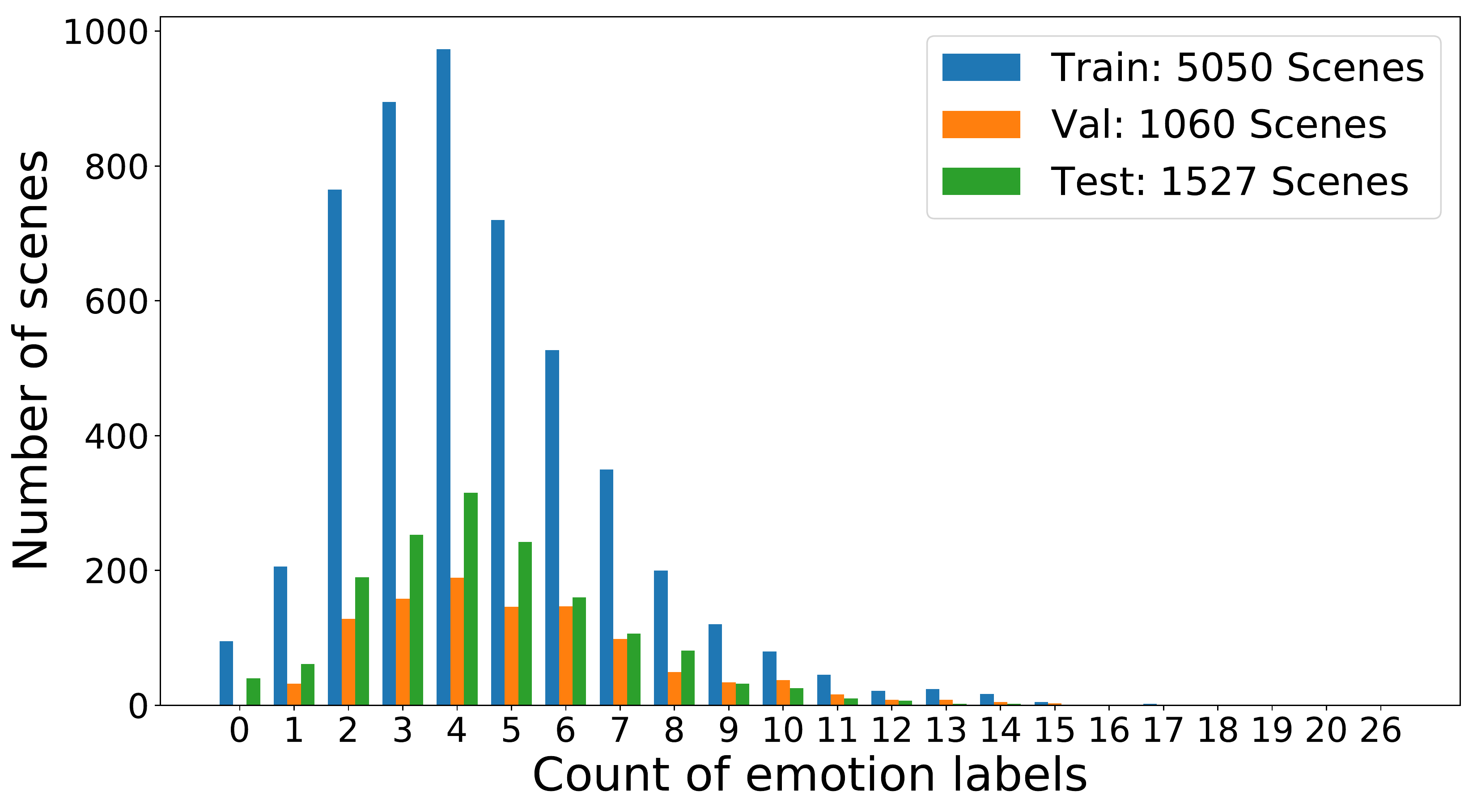}
\vspace{-6mm}
\caption{Number of scenes with a specific number of emotion labels in the train, val, and test splits.}
\vspace{-2mm}
\label{fig:emo_freq}
\end{figure}

Fig.~\ref{fig:emo_freq} shows the number of scenes that have a certain number of emotions.
We observe that most scenes have 2-7 emotions, and the train, val, and test distributions are relatively similar.
The absolute counts are expected to be lower due to smaller val/test sizes.

\begin{figure}[t]
\centering
\includegraphics[width=\linewidth]{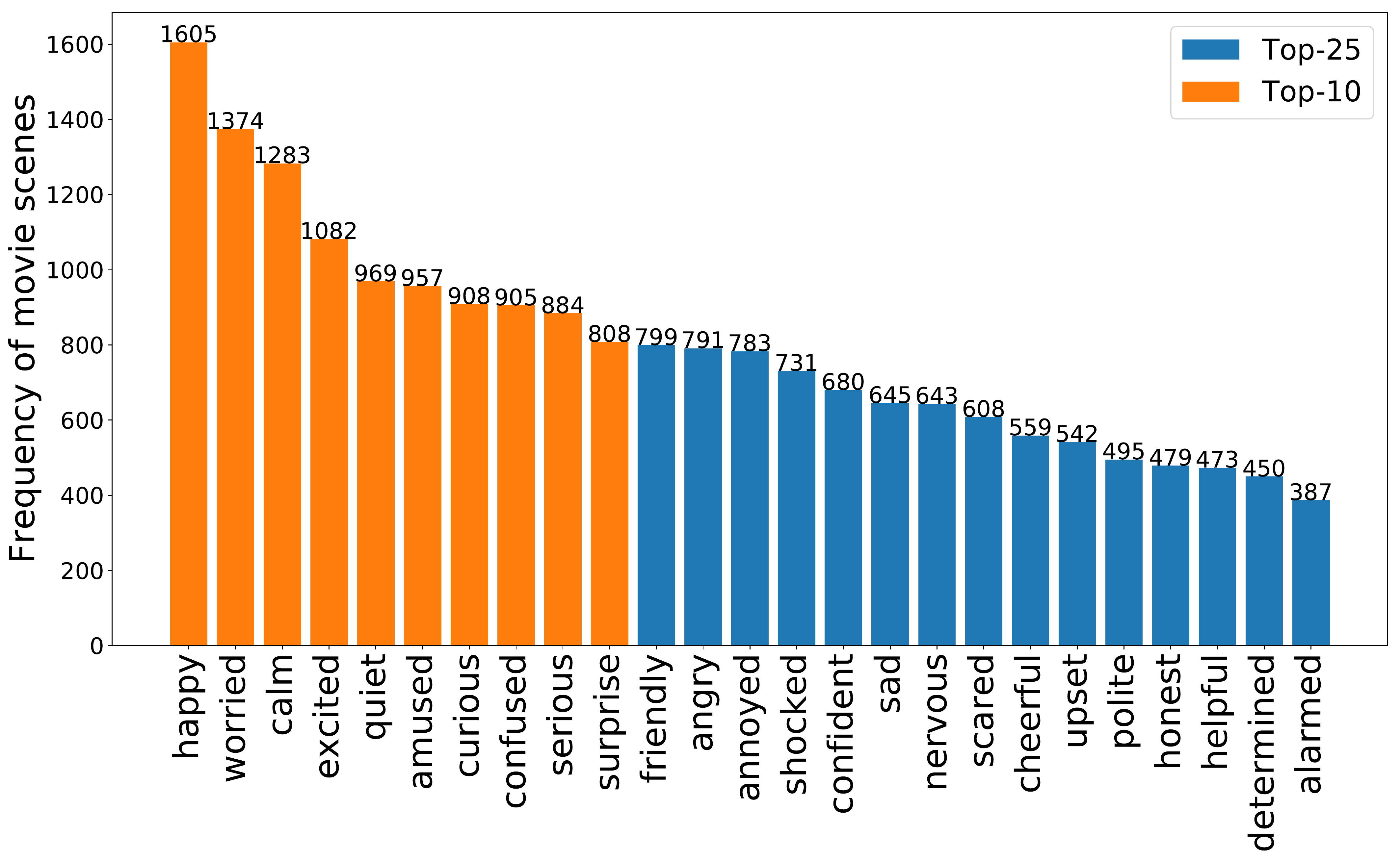}
\vspace{-6mm}
\caption{Number of movie scenes containing top-10 and 25 emotions. Note, the top-25 label set includes top-10.}
\vspace{-2mm}
\label{fig:cg_with_top_10_25_emo}
\end{figure}

Fig.~\ref{fig:cg_with_top_10_25_emo} presents the number of instances for top-10 (orange) and top-25 (orange + blue) label sets.
We see a classic long-tail effect, however, by selecting the top-25, we ensure that there are sufficient instances for all labels to learn a decent representation.

\begin{figure}[t]
\centering
\includegraphics[width=\linewidth]{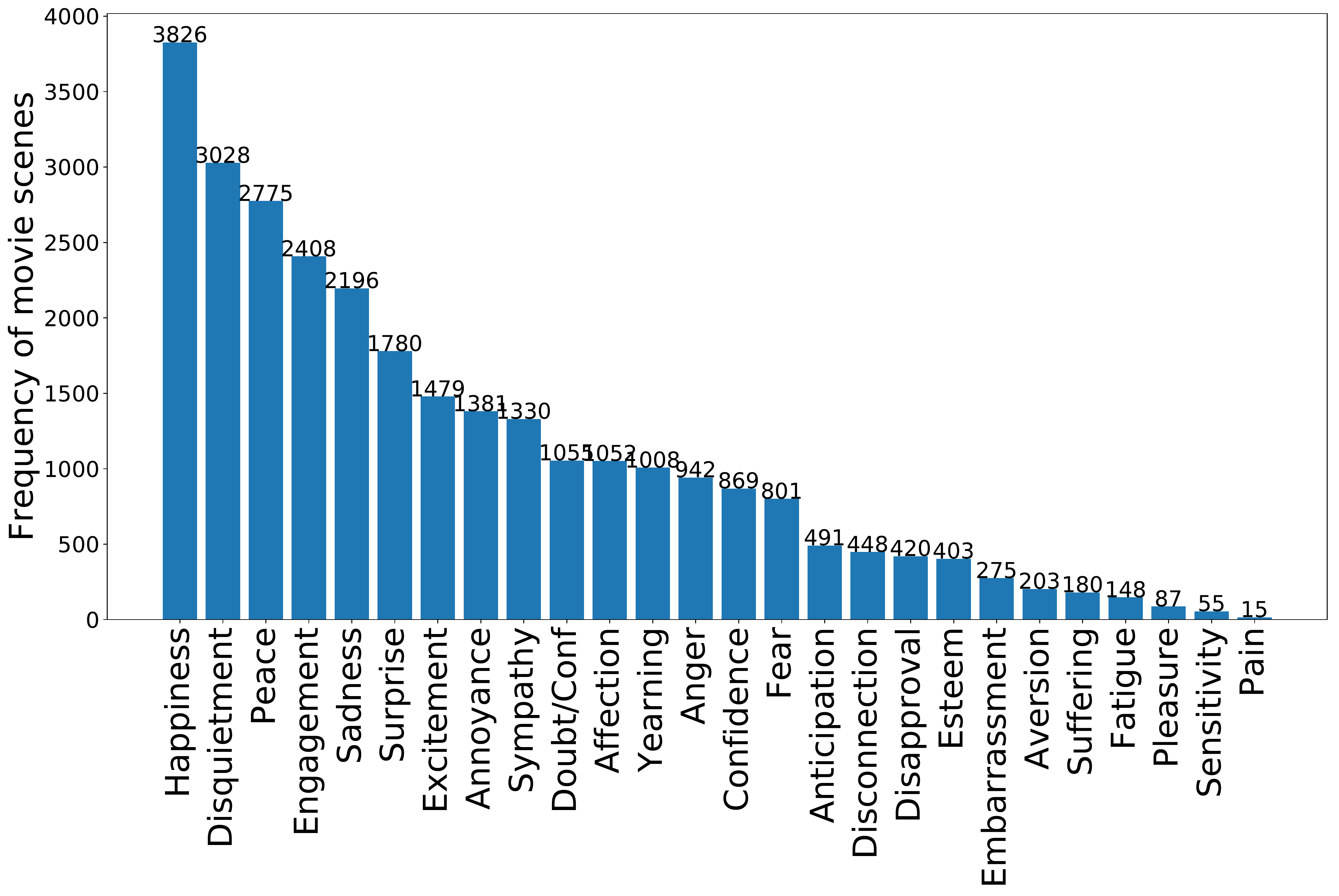}
\vspace{-6mm}
\caption{Number of movie scenes depicting each of the 26 Emotic mapped emotions.}
\vspace{-2mm}
\label{fig:cg_with_emotic_labels}
\end{figure}

Fig.~\ref{fig:cg_with_emotic_labels} shows the same distribution after mapping 181 emotions from MovieGraphs to the 26 emotion labels of the Emotic dataset~\cite{emotic}.
We used a similar mapping as shared by~\cite{affect2mm} and show the details in Table~\ref{tab:emotic_mapping}.
Recall that we report results on this label set in our SoTA experiments in Sec.~\ref{subsec:exp:sota} of the main paper.

\begin{figure}[t]
\centering
\includegraphics[width=\linewidth]{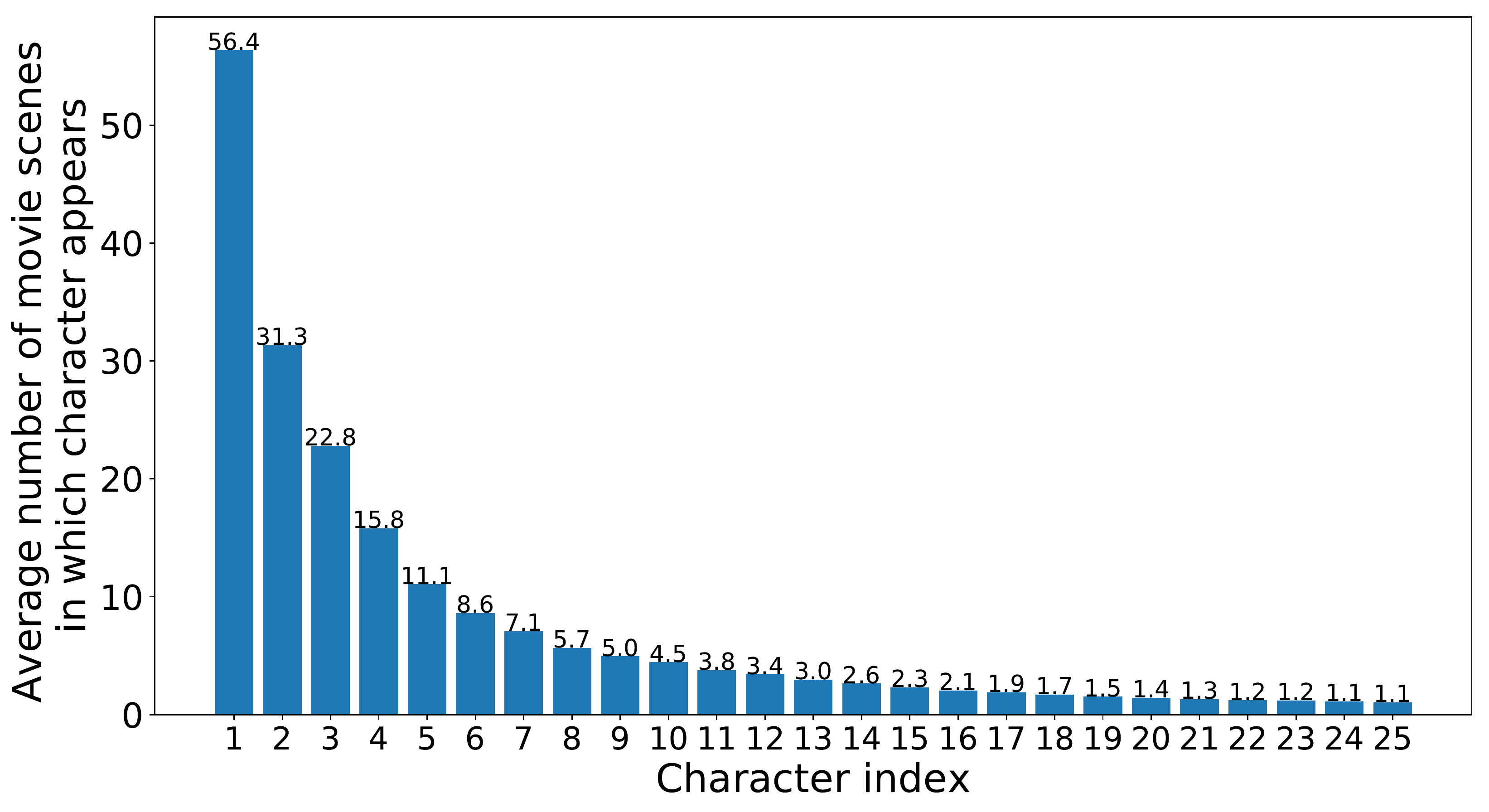}
\vspace{-6mm}
\caption{Average number of scenes in which characters appear. Character 1 corresponds to the most frequently occurring character in the movie, character 2 to the second most frequent, and so on.
For our work, a character is considered as present if there is at least one emotion annotated in the scene.}
\vspace{-2mm}
\label{fig:char_freq}
\end{figure}

We assign the character index 1, 2, $\ldots$ to the most frequent, second most frequent character, and so on. 
The plot in Fig.~\ref{fig:char_freq} shows the average number of scenes in which a character appears, or rather, has an annotated emotion from the MovieGraphs dataset.
This provides interesting avenues for future research, to track emotions across the entire movie.

\begin{figure*}[t]
\centering
\includegraphics[width=0.49\linewidth]{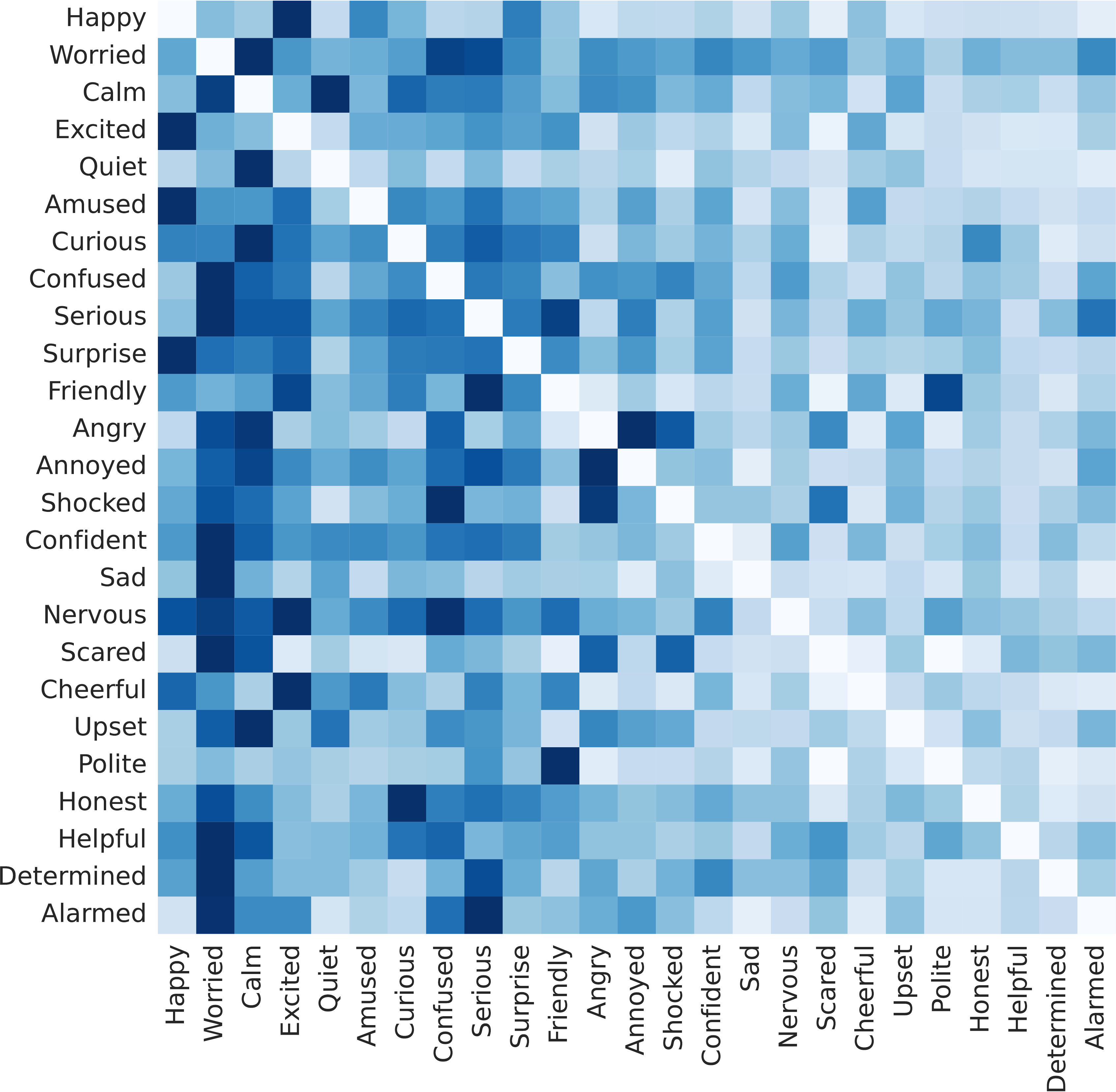}
\includegraphics[width=0.49\linewidth]{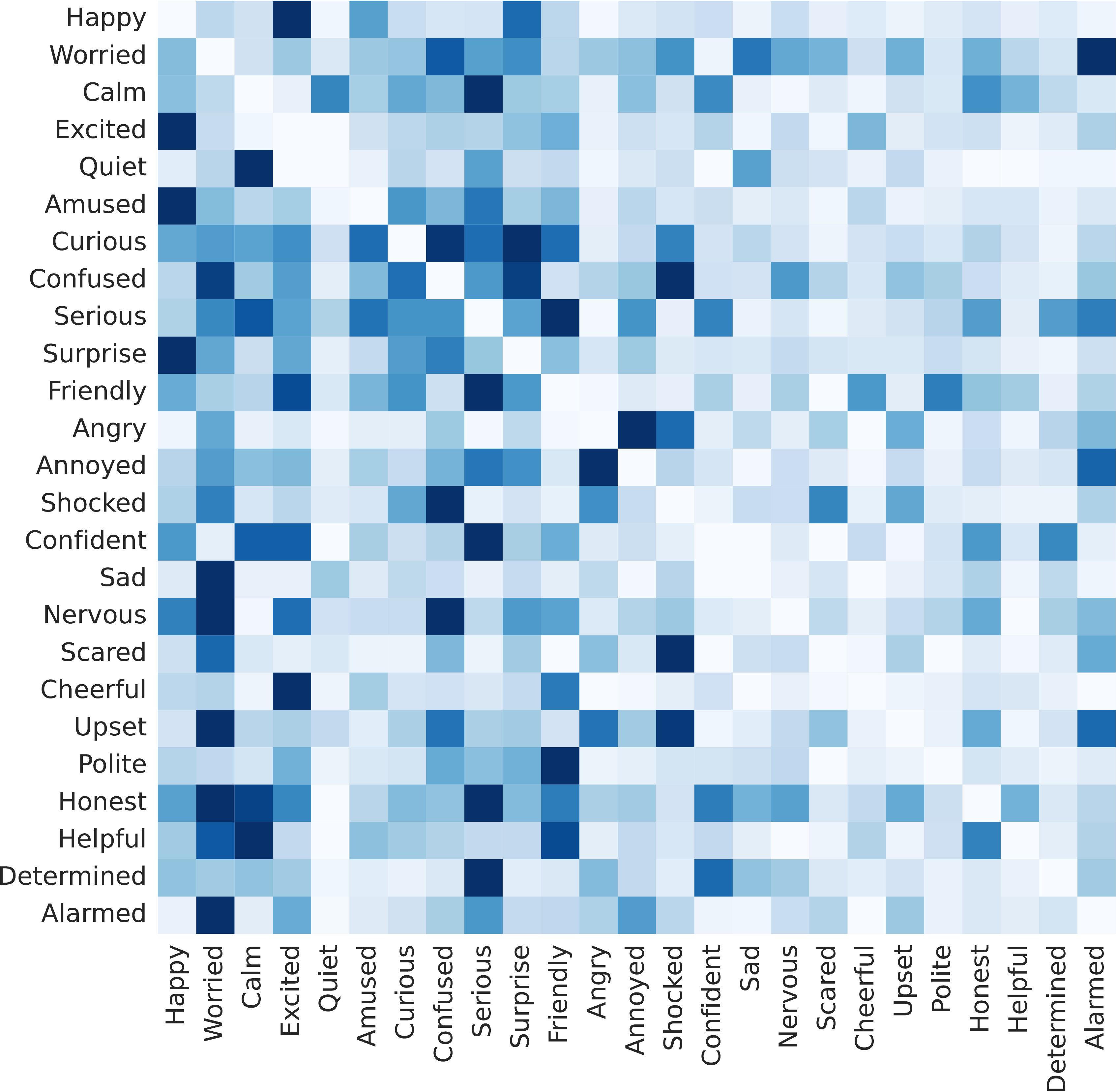}
\vspace{-2mm}
\caption{Normalized label co-occurrence matrices for the top-25 emotions associated with a \emph{movie scene} (left) and \emph{character-level emotions} (right).}
\vspace{-2mm}
\label{fig:cooccurrence_maps_25}
\end{figure*}

\paragraph{Co-occurrence in the top-25 labels.}
Similar to Fig.~\ref{fig:cooccurrence_maps} of the main paper, we show the row-normalized co-occurrence matrices for the top-25 labels in Fig.~\ref{fig:cooccurrence_maps_25}.
From a cursory look, we observe that the movie scene labels (left) are denser than the per-character co-occurrence (right) - this is expected as the movie scene level labels contain a combination of multiple characters.

We present a few notable differences between the scene-level and character-level co-occurrences.
Tuples here correspond to label1 selecting a row, and label2 selecting a column.
(\emph{friendly}, \emph{polite}) seems to be applicable to different characters in a scene, but not for one.
(\emph{honest}, \emph{curious}) shows similar characteristics.
Interestingly while a single character is (\emph{alarmed}, \emph{worried}), in a scene, (\emph{alarmed}, \emph{serious}) also gets fairly high scores.

\begin{table}[t]
\centering
\footnotesize
\tabcolsep=0.05cm
\begin{tabular}{cp{6.2cm}}
\toprule
Emotic Label  & MovieGraphs emotions                                                                                                                                                  \\ \midrule
Affection     & loving, friendly                                                                                                                                                      \\
Anger         & angry, resentful, outraged, vengeful                                                                                                                                  \\
Annoyance     & annoyed, annoying, frustrated, irritated, agitated, bitter, insensitive, exasperated, displeased                                                                      \\
Anticipation  & optimistic, hopeful, imaginative, eager                                                                                                                               \\
Aversion      & disgusted, horrified, hateful                                                                                                                                         \\
Confidence    & confident, proud, stubborn, defiant, independent, convincing                                                                                                          \\
Disapproval   & disapproving, hostile, unfriendly, mean, disrespectful, mocking, condescending, cunning, manipulative, nasty, deceitful, conceited, sleazy, greedy, rebellious, petty \\
Disconnection & indifferent, bored, distracted, distant, uninterested, self-centered, lonely, cynical, restrained, unimpressed, dismissive                                            \\
Disquietment  & worried, nervous, tense, anxious, afraid, alarmed, suspicious, uncomfortable, hesitant, reluctant, insecure, stressed, unsatisfied, solemn, submissive                \\
Doubt/Conf    & confused, skeptical, indecisive                                                                                                                                       \\
Embarrassment & embarrassed, ashamed, humiliated                                                                                                                                      \\
Engagement    & curious, serious, intrigued, persistent, interested, attentive, fascinated                                                                                            \\
Esteem        & respectful, grateful                                                                                                                                                  \\
Excitement    & excited, enthusiastic, energetic, playful, impatient, panicky, impulsive, hasty                                                                                       \\
Fatigue       & tired, sleepy, dizzy                                                                                                                                                  \\
Fear          & scared, fearful, timid, terrified                                                                                                                                     \\
Happiness     & cheerful, delighted, happy, amused, laughing, thrilled, smiling, pleased, overwhelmed, ecstatic, exuberant                                                            \\
Pain          & hurt                                                                                                                                                                  \\
Peace         & content, relieved, relaxed, calm, quiet, satisfied, reserved, carefree                                                                                                \\
Pleasure      & funny, attracted, aroused, hedonistic, pleasant, flattered, entertaining, mesmerized                                                                                  \\
Sadness       & sad, melancholy, upset, disappointed, discouraged, grumpy, crying, regretful, grief-stricken, depressed, heartbroken, remorseful, hopeless, pensive, miserable        \\
Sensitivity   & apologetic, nostalgic                                                                                                                                                 \\
Suffering     & offended, insulted, ignorant, disturbed, abusive, offensive                                                                                                           \\
Surprise      & surprise, surprised, shocked, amazed, startled, astonished, speechless, disbelieving, incredulous                                                                     \\
Sympathy      & kind, compassionate, supportive, sympathetic, encouraging, thoughtful, understanding, generous, concerned, dependable, caring, forgiving, reassuring, gentle          \\
Yearning      & jealous, determined, aggressive, desperate, focused, dedicated, diligent                                                                                              \\ \bottomrule
\end{tabular}
\caption{Mapping MovieGraphs emotions to Emotic labels, adapted from Affect2MM~\cite{affect2mm}.}
\vspace{-4mm}
\label{tab:emotic_mapping}
\end{table}

\section{Character Processing Pipeline}
\label{sec:char_track_extension}

The face tracks provided by the MovieGraphs dataset~\cite{moviegraphs} occasionally miss the characters due to the quality of the face detection.
By watching some clips, we observed that many face tracks were broken within a clip due to missed detections and multiple track IDs were provided for the same character within a single shot.
In addition, some shots had 0 detections, but could be useful to provide a wider perspective on the emotions of that character and scene.
Therefore, we extend the face tracks from MovieGraph dataset by first extending the sparse ground-truth tracks within a shot and then over multiple shots within a scene through clustering.

\begin{figure}[t]
\centering
\includegraphics[width=\linewidth]{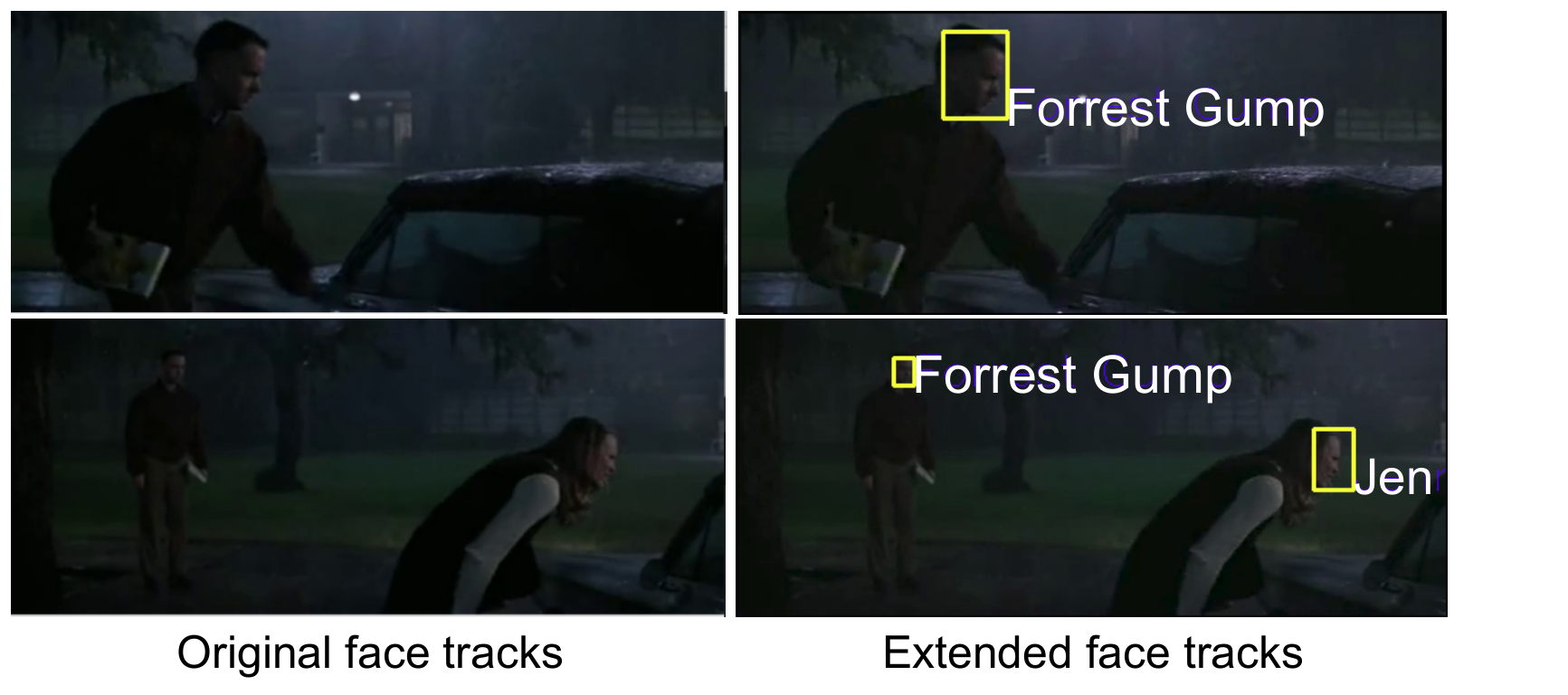}
\vspace{-6mm}
\caption{Example face detections. The original face tracks do not work for dark scenes or profile faces, while our new detections and tracks are able to find them. Scene-036 from Forrest Gump, 1994.}
\vspace{-2mm}
\label{fig:c1c_results}
\end{figure}

In summary, we first recompute face detections and tracks for the movie scenes.
A subset of the new face tracks are assigned a name based on overlap with the original tracks present in the dataset.
Then, we cluster all detections in a clip using hierarchical clustering and assign names to remaining unnamed tracks based on the clustering.
Fig.~\ref{fig:c1c_results} shows an example where original tracks did not have a single detection (due to the dark scene) for a scene in the ``Forrest Gump, 1994" movie.

\paragraph{Face and person detection and tracking.}
\label{subsec:face_tracking}
New face and person detections are extracted from every movie scene of the MovieGraphs dataset.
We adopt MTCNN (Multi-Task Cascaded Convolutional Neural Networks)~\cite{mtcnn} for face detection and Cascade-RCNN pretrained on cast annotations of MovieNet~\cite{movienet} for person detection.
Since the original tracks are only for faces, we first compute person boxes using the person detector and obtain face detections within the person box in order to define a mapping between face and person detections.
If multiple faces are found within a person bounding box, the face with higher detection probability is selected.
The resulting bounding boxes are tracked using the Kalman-filter based SORT (Simple Online and Realtime Tracking) algorithm~\cite{sort}. 
Due to the mapping established between the face and person detections, the same track ID is shared between face and person tracks.
For the rest of the discussion, we focus on face tracks.

\paragraph{Extending names from original to new face tracks.}
Since some of the newly generated tracks coincide with the original tracks from MovieGraphs, such tracks are assigned a name based on their IoU overlap score.
In particular, for every detection in the original tracks, a corresponding new detection is mapped if the IoU score between the two is greater than a threshold (0.7 in our case).
Thus, names from the original detections (or track), are mapped to the new track, and a majority vote of these names is used to decide the final name for a new track.

\paragraph{Face clustering and naming other tracks.}
Not all tracks are assigned a name through the above method due to missed detections in the original tracks.
Thus, we perform clustering to increase the coverage.
First, we extract good identity features from an InceptionResNetV1~\cite{SzegedyInceptionNet2015} pretrained on the VGGFace2~\cite{CaoVGGF22018} dataset.
For clustering we use the C1C~\cite{c1c} algorithm which also uses track information for establishing must and cannot links between the face features.
Individual face detections (features) are processed and clustered using C1C resulting in multiple partitions with varying number of clusters.
We calculate the Silhouette score~\cite{RousseeuwSilhoutte1987} for every partition and the one with highest score is selected as the representative partition.
Now, based on the named tracks generated using the paragraph above, every cluster is assigned a probability corresponding to distinct names (via named detections) within the cluster. 
For clusters which do not have any named detection, equal probability is given to every name present in the scene.
The cluster name-probabilities corresponding to the detections of unnamed tracks are extracted and the average of these soft scores is used to reflect the names for the newly discovered tracks.
This way, we assign a name probability to new tracks and threshold it with 0.7 to select the final name for such new tracks.

\section{Analyzing AP scores}
\label{sec:AP_analysis}

\begin{figure*}[t]
\centering
\includegraphics[width=0.8\linewidth]{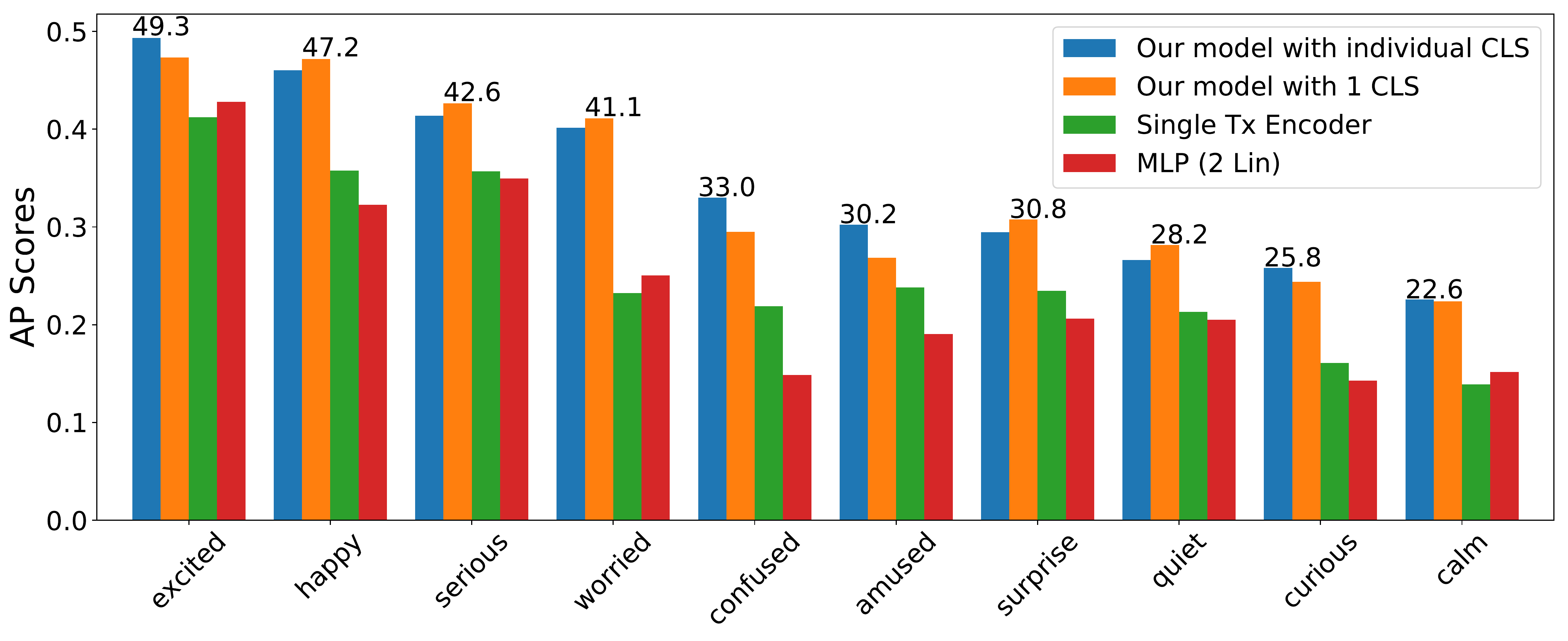}
\vspace{-4mm}
\caption{AP scores for the top-10 emotions label set sorted from high to low AP score for our model with individual CLS tokens.}
\label{fig:t10_AP}
\vspace{4mm}
\includegraphics[width=0.8\linewidth]{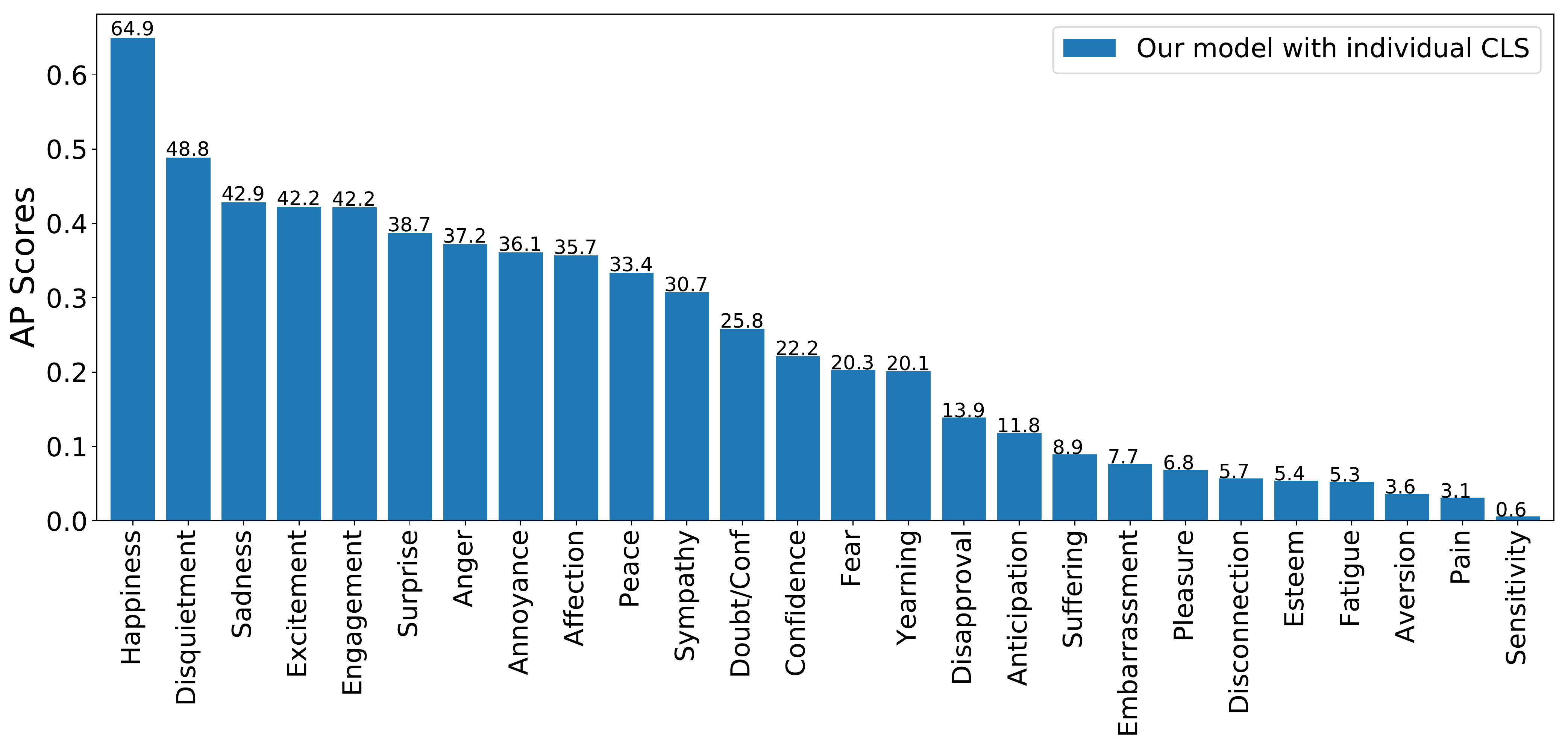}
\vspace{-4mm}
\caption{AP scores on the 26 grouped labels of the Emotic label set.}
\label{fig:emotic_AP}
\end{figure*}

Similar to Fig.~\ref{fig:per_class_ap} of the main paper, we present per-emotion scores for the top-10 emotions in the dataset in Fig.~\ref{fig:t10_AP}.
We observe that our model with the individual classifier (CLS) tokens outperforms other approaches in 5 of 10 emotions.
In Fig.~\ref{fig:emotic_AP}, we show the AP for each group of Emotic labels.
We observe that challenging labels such as \emph{pain}, \emph{sensitivity}, perform much worse than others such as \emph{happiness}, \emph{sadness}, \emph{anger}, \etc.

\section{Feature Ablation}
\label{sec:feature_abl}

\begin{table*}[t]
\centering
\tabcolsep=0.18cm
\begin{tabular}{l ccc | ccc | cc | cc cc}
\toprule
& \multicolumn{3}{c|}{Video} & \multicolumn{3}{c|}{Character} & \multicolumn{2}{c|}{Dialog} & \multicolumn{4}{c}{Metrics (mAP)} \\
& \multicolumn{1}{c}{MViT} & \multicolumn{1}{c}{R50} & \multicolumn{1}{c|}{R152} & \multicolumn{1}{c}{R50} & \multicolumn{1}{c}{VGG-M} & \multicolumn{1}{c|}{IRv1} & \multicolumn{1}{c}{RB} & \multicolumn{1}{c|}{RB} & \multicolumn{2}{c}{Top-10} & \multicolumn{2}{c}{Top-25} \\
& \multicolumn{1}{c}{K400} & \multicolumn{1}{c}{P365} & \multicolumn{1}{c|}{INet} & \multicolumn{1}{c}{FER} & \multicolumn{1}{c}{FER} & \multicolumn{1}{c|}{VGG-F} & \multicolumn{1}{c}{FT} & \multicolumn{1}{c|}{PT} & \multicolumn{1}{c}{Scene} & \multicolumn{1}{c}{Char} & \multicolumn{1}{c}{Scene} & \multicolumn{1}{c}{Char} \\ \midrule
1 & - & \dingcheck & - & - & - & \dingcheck & - & \dingcheck & 25.07\scriptsize{{$\pm$}0.12} & 15.48\scriptsize{{$\pm$}0.15} & 16.41\scriptsize{{$\pm$}0.24} & 8.31\scriptsize{{$\pm$}0.17} \\
2 & - & - & \dingcheck & - & - & \dingcheck & - & \dingcheck & 25.85\scriptsize{{$\pm$}0.24} & 15.63\scriptsize{{$\pm$}0.21} & 16.45\scriptsize{{$\pm$}0.09} & 8.31\scriptsize{{$\pm$}0.09} \\
3 & - & - & \dingcheck & - & \dingcheck & - & - & \dingcheck & 29.20\scriptsize{{$\pm$}0.22} & 19.88\scriptsize{{$\pm$}0.27} & 18.93\scriptsize{{$\pm$}0.38} & 10.16\scriptsize{{$\pm$}0.17} \\
4 & \dingcheck & - & - & - & - & \dingcheck & - & \dingcheck & 29.27\scriptsize{{$\pm$}0.08} & 18.07\scriptsize{{$\pm$}0.22} & 18.35\scriptsize{{$\pm$}0.09} & 0.09\scriptsize{{$\pm$}0.08} \\
5 & - & \dingcheck & - & - & \dingcheck & - & - & \dingcheck & 29.30\scriptsize{{$\pm$}0.21} & 19.73\scriptsize{{$\pm$}0.17} & 19.05\scriptsize{{$\pm$}0.19} & 10.31\scriptsize{{$\pm$}0.00} \\
6 & \dingcheck & - & - & - & \dingcheck & - & - & \dingcheck & 29.34\scriptsize{{$\pm$}0.08} & 20.50\scriptsize{{$\pm$}0.04} & 19.07\scriptsize{{$\pm$}0.19} & 10.34\scriptsize{{$\pm$}0.17} \\
7 & - & \dingcheck & - & - & - & \dingcheck & \dingcheck & - & 29.34\scriptsize{{$\pm$}0.17} & 19.49\scriptsize{{$\pm$}0.03} & 20.73\scriptsize{{$\pm$}0.08} & 10.75\scriptsize{{$\pm$}0.02} \\
8 & - & - & \dingcheck & - & - & \dingcheck & \dingcheck & - & 29.47\scriptsize{{$\pm$}0.14} & 19.29\scriptsize{{$\pm$}0.10} & 20.74\scriptsize{{$\pm$}0.11} & 10.79\scriptsize{{$\pm$}0.07} \\
9 & - & \dingcheck & - & \dingcheck & - & - & - & \dingcheck & 29.69\scriptsize{{$\pm$}0.38} & 20.25\scriptsize{{$\pm$}0.14} & 20.16\scriptsize{{$\pm$}0.29} & 11.06\scriptsize{{$\pm$}0.12} \\
10 & - & - & \dingcheck & \dingcheck & - & - & - & \dingcheck & 30.19\scriptsize{{$\pm$}0.38} & 20.27\scriptsize{{$\pm$}0.26} & 19.83\scriptsize{{$\pm$}0.07} & 11.06\scriptsize{{$\pm$}0.16} \\
11 & \dingcheck & - & - & \dingcheck & - & - & - & \dingcheck & 31.39\scriptsize{{$\pm$}0.34} & 21.18\scriptsize{{$\pm$}0.18} & 20.88\scriptsize{{$\pm$}0.28} & 11.46\scriptsize{{$\pm$}0.08} \\
12 & \dingcheck & - & - & - & \dingcheck & - & \dingcheck & - & 31.50\scriptsize{{$\pm$}0.36} & 21.60\scriptsize{{$\pm$}0.09} & 21.49\scriptsize{{$\pm$}0.30} & 11.64\scriptsize{{$\pm$}0.20} \\
13 & - & - & \dingcheck & - & \dingcheck & - & \dingcheck & - & 31.96\scriptsize{{$\pm$}0.20} & 21.81\scriptsize{{$\pm$}0.37} & 21.28\scriptsize{{$\pm$}0.25} & 11.58\scriptsize{{$\pm$}0.26} \\
14 & \dingcheck & - & - & - & - & \dingcheck & \dingcheck & - & 32.23\scriptsize{{$\pm$}0.07} & 21.45\scriptsize{{$\pm$}0.07} & 22.10\scriptsize{{$\pm$}0.11} & 11.63\scriptsize{{$\pm$}0.06} \\
15 & - & \dingcheck & - & - & \dingcheck & - & \dingcheck & - & 32.42\scriptsize{{$\pm$}0.26} & 22.32\scriptsize{{$\pm$}0.27} & 21.45\scriptsize{{$\pm$}0.17} & 11.62\scriptsize{{$\pm$}0.05} \\
16 & - & - & \dingcheck & \dingcheck & - & - & \dingcheck & - & 33.44\scriptsize{{$\pm$}0.33} & 22.89\scriptsize{{$\pm$}0.24} & 22.75\scriptsize{{$\pm$}0.18} & 12.52\scriptsize{{$\pm$}0.12} \\
17 & - & \dingcheck & - & \dingcheck & - & - & \dingcheck & - & 33.46\scriptsize{{$\pm$}0.21} & 22.98\scriptsize{{$\pm$}0.16} & 22.69\scriptsize{{$\pm$}0.22} & 12.48\scriptsize{{$\pm$}0.20} \\
18 & \dingcheck & - & - & \dingcheck & - & - & \dingcheck & - & \textbf{34.22}\scriptsize{{$\pm$}0.18} & \textbf{24.35}\scriptsize{{$\pm$}0.23} & \textbf{23.86}\scriptsize{{$\pm$}0.10} & \textbf{13.36}\scriptsize{{$\pm$}0.11} \\ \bottomrule
\end{tabular}
\caption{Extended feature ablations. The different feature backbones are 
(MViT, K400): MViT pretrained on Kinetics400,
(R50, P365): ResNet50 on Places365,
(R152, INet): ResNet152 on ImageNet,
(R50, FER): ResNet50 on Facial Expression Recognition (FER),
(VGG-M, FER): VGG-M on FER,
(IRv1, VGG-F): InceptionResNet-v1 trained on VGG-Face dataset,
(RB, FT): pretrained RoBERTa finetuned for emotion recognition and
(RB, PT): pretrained RoBERTa. Best numbers in bold, close second in italics.}
\label{tab:supp_feat_abl}
\end{table*}

We expand upon the feature ablation in Table~\ref{tab:feat_abl} of the main paper to show the effect of additional feature combinations in Table~\ref{tab:supp_feat_abl}.
All the trends are similar, fine-tuning RoBERTa helps consistently, ResNet50 trained on FER appears to be a good representation for characters, and the MViT trained on Kinetics400 provides better results for both the label sets, while ResNet50 trained on Places365 is a close second.

\begin{table*}[t]
\centering
\tabcolsep=0.07cm
\begin{tabular}{l cc cc cc}
\toprule
\multicolumn{1}{c}{\multirow{2}{*}{Method}} & \multicolumn{2}{c}{Top 10}      & \multicolumn{2}{c}{Top 25}      & \multicolumn{2}{c}{Emotic}      \\
\multicolumn{1}{c}{} & Val   & Test  & Val   & Test  & Val   & Test  \\ \midrule
Random               & 16.87\scriptsize{{$\pm$}0.23} & 13.84\scriptsize{{$\pm$}0.20} & 9.73\scriptsize{{$\pm$}0.10} & 7.57\scriptsize{{$\pm$}0.08} & 11.47\scriptsize{{$\pm$}0.11} & 11.36\scriptsize{{$\pm$}0.09} \\
CAER~\cite{caer}
& 18.35\scriptsize{{$\pm$}0.10} & 15.38\scriptsize{{$\pm$}0.13} & 11.84\scriptsize{{$\pm$}0.07} & 9.49\scriptsize{{$\pm$}0.08}  & 13.91\scriptsize{{$\pm$}0.06} & 12.68\scriptsize{{$\pm$}0.02} \\
ENet~\cite{WeiEmotionNet}
& 19.14\scriptsize{{$\pm$}0.10} & 16.14\scriptsize{{$\pm$}0.05} & 11.22\scriptsize{{$\pm$}0.06} & 9.08\scriptsize{{$\pm$}0.08}  & 13.55\scriptsize{{$\pm$}0.06} & 12.64\scriptsize{{$\pm$}0.03} \\
AANet~\cite{attendaffectnet}
& 21.55\scriptsize{{$\pm$}0.18} & 17.55\scriptsize{{$\pm$}0.16} & 12.55\scriptsize{{$\pm$}0.15} & 10.20\scriptsize{{$\pm$}0.13} & 14.71\scriptsize{{$\pm$}0.19} & 13.37\scriptsize{{$\pm$}0.20} \\
M2Fnet~\cite{m2fnet}
& 24.55\scriptsize{{$\pm$}0.39} & 19.10\scriptsize{{$\pm$}0.06} & 16.02\scriptsize{{$\pm$}0.14} & 13.05\scriptsize{{$\pm$}0.31} & 18.27\scriptsize{{$\pm$}0.16} & 16.76\scriptsize{{$\pm$}0.20} \\
\midrule
\modelname{}  & \textbf{34.22}\scriptsize{{$\pm$}0.18} & \textbf{29.35}\scriptsize{{$\pm$}0.18} & \textbf{23.86}\scriptsize{{$\pm$}0.10} & \textbf{19.47}\scriptsize{{$\pm$}0.10} & \textbf{23.67}\scriptsize{{$\pm$}0.03} & \textbf{21.40}\scriptsize{{$\pm$}0.03} \\ \bottomrule
\end{tabular}
\vspace{-3mm}
\caption{Comparison against SoTA for scene-level predictions. \emph{AANet} denotes AttendAffectNet, while \emph{ENet} refers to EmotionNet.}
\vspace{-2mm}
\label{tab:suppl_sota_scene_abl}
\end{table*}

\begin{table*}[t]
\centering
\tabcolsep=0.09cm
\begin{tabular}{l cc cc cc}
\toprule
\multicolumn{1}{c}{\multirow{2}{*}{Method}} & \multicolumn{2}{c}{Top 10}      & \multicolumn{2}{c}{Top 25}      & \multicolumn{2}{c}{Emotic}      \\
\multicolumn{1}{c}{} & Val   & Test  & Val   & Test & Val   & Test \\
\midrule
Random
& 12.49\scriptsize{{$\pm$}0.15} & 11.37\scriptsize{{$\pm$}0.14} & 5.84\scriptsize{{$\pm$}0.05} & 5.36\scriptsize{{$\pm$}0.05} & 6.40\scriptsize{{$\pm$}0.05} & 6.32\scriptsize{{$\pm$}0.05} \\
AANet~\cite{attendaffectnet}
& 17.43\scriptsize{{$\pm$}0.28} & 16.04\scriptsize{{$\pm$}0.19} & 8.64\scriptsize{{$\pm$}0.19} & 7.20\scriptsize{{$\pm$}0.15} & 8.53\scriptsize{{$\pm$}0.17} & 7.75\scriptsize{{$\pm$}0.11} \\
M2Fnet~\cite{m2fnet}
& 20.82\scriptsize{{$\pm$}0.28} & 19.01\scriptsize{{$\pm$}0.45} & 10.67\scriptsize{{$\pm$}0.38}  & 9.71\scriptsize{{$\pm$}0.34} & 11.30\scriptsize{{$\pm$}0.35}  & 9.92\scriptsize{{$\pm$}0.02} \\
\midrule
\modelname{} (Ours)
& \textbf{24.35}\scriptsize{{$\pm$}0.23} & \textbf{22.31}\scriptsize{{$\pm$}0.11} & \textbf{13.36}\scriptsize{{$\pm$}0.11} & \textbf{11.71}\scriptsize{{$\pm$}0.05} & \textbf{12.29}\scriptsize{{$\pm$}0.08} & \textbf{11.76}\scriptsize{{$\pm$}0.10} \\ \bottomrule
\end{tabular}
\vspace{-2mm}
\caption{Comparison against SoTA for character-level predictions. \emph{AANet} denotes AttendAffectNet.}
\vspace{-5mm}
\label{tab:suppl_char_sota_abl}
\end{table*}

\section{Adapting SoTA Methods for our Task}
\label{sec:sota_adaptations}

\begin{figure*}[t]
\centering
\includegraphics[width=\linewidth]{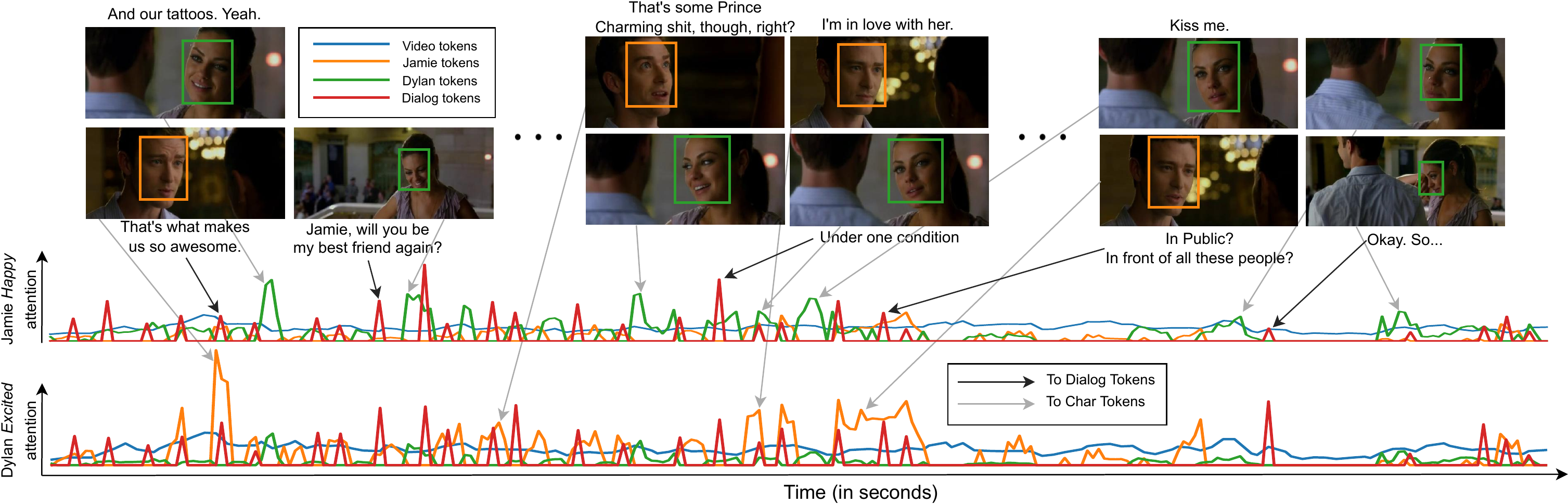}
\caption{A scene from the movie \textit{Friends with Benefits} with self-attention scores for multiple modalities for two character-level predictions: \textit{Jamie is happy} and \textit{Dylan is excited}.
From the figure we can infer that the \textit{happy} classifier token attends to the \textit{Jamie} character tokens with spikes observed when she smiles or laughs, while \textit{Dylan's excited} classifier token attends primarily to the dialog utterance tokens.
We can see this as very few face snaps indicate that \textit{Dylan} is excited, in fact, \emph{Dylan's} face is not even visible often.
However, dialog utterances like \textit{That's what makes us so awesome}, \textit{Hey, I miss you}, and \textit{Jamie, will you be my best friend again?} are extremely useful for the model to infer the emotions.}
\label{fig:fwb_qa}
\vspace{-5mm}
\end{figure*}

The MovieGraphs dataset has not been used directly to predict emotions at a scene or character level.
Related to using labels from MovieGraphs, Affect2MM~\cite{affect2mm} extracts scene-level emotion timelines for the entire movie, but relies on one emotion per scene.
This is quite different from our vision of a multi-label setting where the scene and each character can present multiple emotions.
For a fair comparison to previous work, we chose models that have attained SoTA in image, video and multimodal emotion recognition.
We share details on how these methods are adapted to make them suitable for our task.

\paragraph{EmotionNet}~\cite{WeiEmotionNet}
is a recent SoTA for emotion recognition from web images.
It uses a joint embedding training approach which uses emotional keywords associated with a given image and aligns its learned text embedding (pretrained on massive text data) with image embedding extracted from a standard feature backbone (ResNet50). 
To adapt EmotionNet for our task, we used word2vec~\cite{MikolovWord2Vec} for extracting text embeddings and ResNet50 for frames.
Since we use a video as input, the frame features are max-pooled to generate a single representation.
We use the proposed embedding loss and provide the emotion labels as the keywords for joint embedding training. This learned ResNet50 is finetuned for multilabel emotion recognition where the individual frame features are max-pooled before passing to the logits layer.

\paragraph{CAER (Context Aware Emotion Recognition)}~\cite{caer}
is a deep Convolutional Network which consists of two stream encoding networks to separately extract the facial and context features which are fused using an adaptive fusion network.
Detections from our extended face tracks are used as inputs for the face encoding stream and the full video frame with masked faces was used as input to context encoding stream.
Since CAER is designed to extract emotions from images we adapt it to videos by applying max-pooling over the fused features from both the streams to generate a single representation for a video.
This adapted model is trained to predict multiple scene-level emotions.

\paragraph{M2FNet}~\cite{m2fnet}
is a transformer based model originally developed for Emotion Recognition in Conversations (ERC) and features a fusion-attention mechanism to modulate the attention given to each utterance considering the audio and visual features.
As this model is designed for utterance emotion recognition we apply a max-pooling operation over the final outputs of fusion attention module to generate a feature representation for all the utterances in a video.
Since this model provides two strategies to consider visual features: one with the video frame and another that combines multiple faces in a frame, we use them to predict either scene- or character-level emotions separately.

\paragraph{AttendAffectNet}~\cite{attendaffectnet}
proposes two multi-modal self-attention based approaches for predicting emotions from movie clips. We adapted the proposed Feature AttendAffectNet model in our work.
It leverages the transformer encoder block where every input token represents a different modality. These modality feature vectors are generated by average pooling over respective features. Following the proposed mechanism, a classification head was attached at the end of the model for predicting multi-label emotions.
We adopt the same backbone representations, MViT~\cite{FanMViT2021} pre-trained on Kinetics400~\cite{CarreiraQuoVadis2017} and ResNet50 pretrained on FER13~\cite{fer13}, for their work to extract scene and face features respectively.

\paragraph{SoTA results.}
Reflecting Tables~\ref{tab:sota_scene_abl} and \ref{tab:char_sota_abl} in the main paper, we present the Table~\ref{tab:suppl_sota_scene_abl} and Table~\ref{tab:suppl_char_sota_abl} and also include standard deviation over 3 runs.

\section{Additional Qualitative Analysis}
\label{sec:supp_qualitative}

Fig.~\ref{fig:fwb_qa} shows another example (similar to Fig.~\ref{fig:qualitative_example} from the main paper) where we visualize the emotions for two characters \emph{Jamie} and \emph{Dylan}.
We see that our model looks at relevant video frames, dialog utterances, and character representations while making the predictions.
The scene described above is of a \textit{proposal}, where the protagonist, \emph{Dylan}, clears out some misunderstanding and proposes to the female lead character, \emph{Jamie}, in between an ongoing flash mob (\textit{scene}).
As mentioned, in the Fig.~\ref{fig:fwb_qa} caption, both the characters develop emotion: \textit{happy} and \textit{excited}.
From the facial expressions as well as from dialog utterances, it is apparent enough for the readers to predict emotions, but from model's point-of-view culminating all these signals and making sense of them, that too for complex human emotions, is a great job.

\paragraph{User study on understanding expressiveness.}
We asked 2 people to look at about 30 random clips that have positive labels for \emph{angry}, \emph{scared}, \emph{cheerful} and independently mark yes when the emotion was apparent in the video (V), dialog (D), and character/face (C), similar to a multi-label setup.
Note, our model's attention scores suggest that \emph{cheerful} is an expressive emotion (character tokens are helpful), while \emph{scared} and \emph{angry} can rely on dialog and video context.

Below, we present the fraction of times each modality was picked by the users.
For \emph{angry}, the annotators favored V: 62\%, D: 80\%, and C: 59\%, due to several neutral-faced instances with harsh dialog and violent actions.
\emph{Scared}, V: 56\%, D: 48\%, C: 62\%, was sometimes expressed through screaming or crying, with no modality standing out strongly.
Finally, \emph{cheerful}, V: 41\%, D: 64\%, C: 79\%, was observed most prominently on character faces and through dialog.
Note that this analysis aligns with our observations in Fig.~\ref{fig:expression} of the main paper that the expressiveness scores are applicable to our particular dataset.

\clearpage

\balance
{\small
\bibliographystyle{ieee_fullname}
\bibliography{longstrings,main}
}

\end{document}